\documentclass{article}

\PassOptionsToPackage{numbers, compress}{natbib}

\usepackage[preprint]{neurips_2026}

\usepackage[utf8]{inputenc}
\usepackage[T1]{fontenc}
\usepackage{hyperref}
\usepackage{url}
\usepackage{booktabs}
\usepackage{amsfonts}
\usepackage{nicefrac}
\usepackage{microtype}
\usepackage{xcolor}

\usepackage[utf8]{inputenc}
\usepackage[T1]{fontenc}
\usepackage{hyperref}
\usepackage{url}
\usepackage{booktabs}
\usepackage{amsfonts}
\usepackage{nicefrac}
\usepackage{microtype}
\usepackage{xcolor}
\usepackage{graphicx}
\usepackage{multirow}
\usepackage{makecell}
\usepackage{array}
\usepackage{tabularx}
\usepackage{ragged2e}
\usepackage{enumitem}
\usepackage{subcaption}
\usepackage{float}
\usepackage{amsmath, amssymb}
\usepackage{bm}
\usepackage{siunitx}
\usepackage{pifont}
\newcommand{\cmark}{\ding{51}}
\newcommand{\xmark}{\ding{55}}
\newcommand{\pmark}{\ding{108}}
\usepackage{arydshln}
\usepackage{placeins}
\usepackage{colortbl}

\newcommand{\dataset}{MMS-VPR}
\newcommand{\library}{MMS-VPRlib}

\title{MMS-VPR: A Fine-Grained Multimodal Street-Level Visual Place Recognition Dataset and Evaluation Benchmark for Dense Pedestrian Environments}

\author{
	Yiwei Ou$^{1,*}$,\,
	Xiaobin Ren$^{2}$,\,
	Ronggui Sun$^{1}$,\,
	Guansong Gao$^{1}$,\,
	Kaiqi Zhao$^{3}$,\,
	Manfredo Manfredini$^{1}$ \\[3pt]
	$^{1}$University of Auckland \quad
	$^{2}$Peking University \quad
	$^{3}$Harbin Institute of Technology, Shenzhen \\[2pt]
	$^{*}$Corresponding author: \texttt{ou.yiwei@auckland.ac.nz}
}

\begin{document}
	\maketitle

	\begin{abstract}
		Existing visual place recognition (VPR) datasets predominantly rely on vehicle-mounted imagery,
		offer limited multimodal diversity, and underrepresent dense pedestrian street scenes,
		particularly in non-Western urban contexts.
		We present \textbf{MMS-VPR}, a challenging fine-grained multimodal benchmark for
		pedestrian-level place recognition in dense urban environments.
		MMS-VPR comprises 110,529 images and 2,527 video clips across \textbf{208 fine-grained
			location classes} within a $\sim$70,800\,m\textsuperscript{2} compact urban district
		in Chengdu, China---with adjacent classes separated by as little as 10--20\,m.
		Unlike large-area VPR datasets, MMS-VPR elevates difficulty through visual ambiguity:
		repetitive storefronts, dense occlusion, and rapid viewpoint changes that challenge both
		CNN-based and Transformer-based state-of-the-art methods.
		Field data were collected in 2024 following a systematic multi-direction, dual-perspective
		protocol; social media imagery spans seven years (2019--2025), providing long-term
		appearance change across 31,726 geolocated photographs.
		Each location features comprehensive day--night coverage, multiple viewing angles, and
		multimodal annotations including GPS coordinates, timestamps, semantic text, and an explicit
		spatial graph with space syntax enrichments.
		We release \textbf{MMS-VPRlib}, a unified benchmarking platform consolidating 22 baselines
		spanning shallow ML, CNN, Transformer, graph neural network (GNN), and multimodal
		architectures under a standardized, reproducible pipeline.
		Controlled modality ablation studies confirm that video temporal features are the most
		powerful single modality (+15.8\,\% accuracy over image-only), semantic text provides
		superior coarse retrieval (R@5 = 0.988 alone), and full multimodal fusion achieves
		0.981 accuracy---a +19.2\,\% gain over image-only---while graph-structured modeling
		provides additional signal in fine-grained structural subsets.
		All data, code, and Croissant metadata are publicly available.
		\footnote{Dataset: \url{huggingface.co/datasets/Yiwei-Ou/MMS-VPR};
			Benchmark: \url{github.com/yiasun/MMS-VPRlib}}
	\end{abstract}

	\section{Introduction}
	\label{sec:intro}

\begin{figure*}[t]
		\centering
		\includegraphics[width=\textwidth]{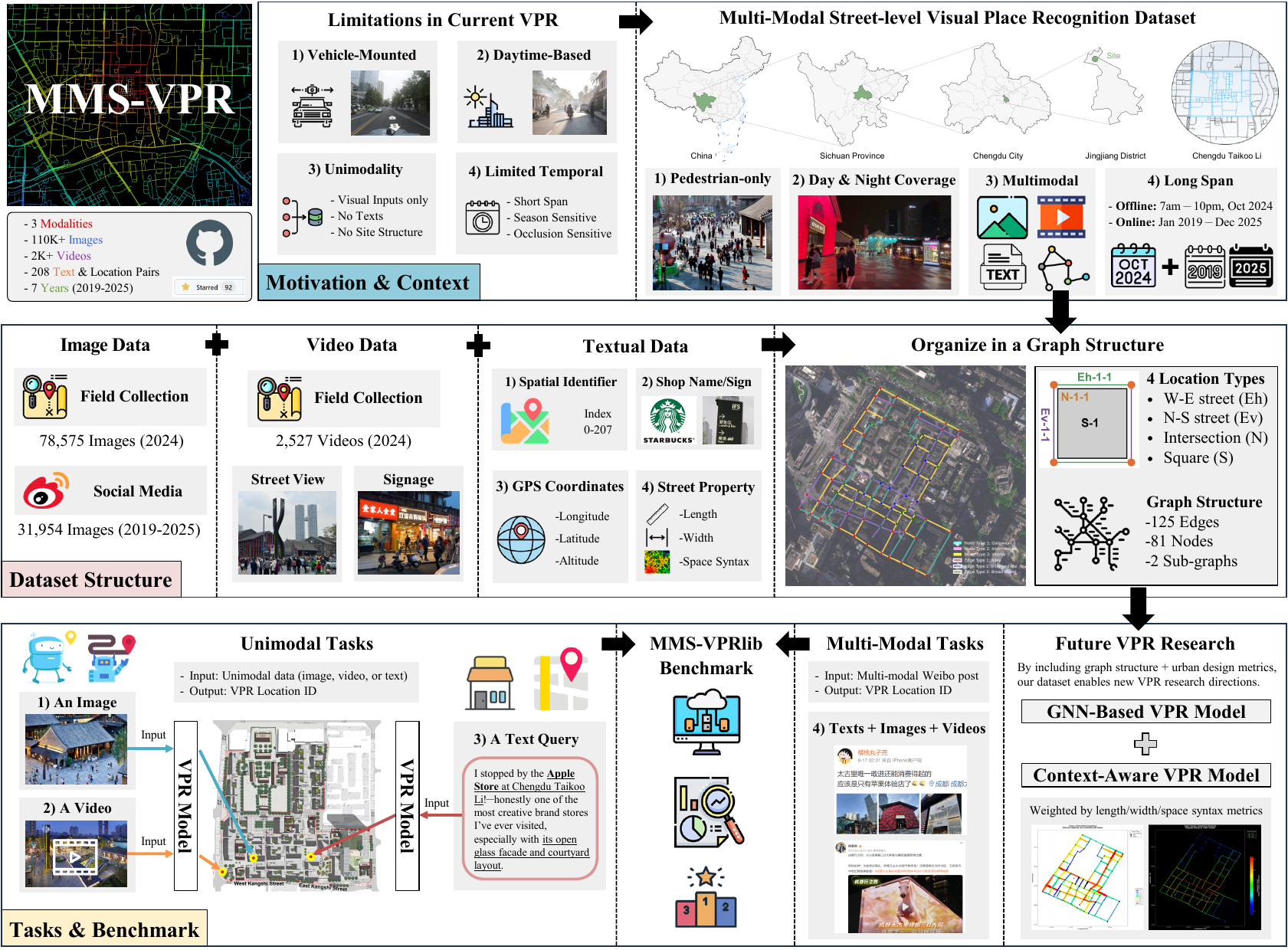}
		\caption{Overview of MMS-VPR framework addressing four VPR limitations through pedestrian-only
			urban environments, day--night coverage, multimodal integration, and 7-year temporal span.}
		\label{fig:framework}
	\end{figure*}

Visual Place Recognition (VPR), also known as visual geolocalization, retrieves visually similar places from a geo-tagged reference database to estimate the location of a query image~\cite{FOL,lowry2015visual}. VPR is central to robotics, augmented reality, urban computing, autonomous navigation, and geospatial AI~\cite{quach2024ms,wu2024gicnet,huang2024dino,FOL}. However, widely used VPR benchmarks still make evaluation easier in several ways that matter for deployment, leaving four systematic gaps that limit applicability to real-world pedestrian environments (Figure~\ref{fig:framework}): \textbf{(1) Vehicle-mounted perspective.}
	Most benchmarks rely on vehicle-mounted cameras (Google Street View, Mapillary)~\cite{CV_Cities,GSV_Cities}, excluding pedestrian-only spaces inaccessible to
	vehicles.
	Studies that do target pedestrian contexts cover either sparsely distributed urban
	areas~\cite{NYC_Event_VPR,brizi2024vbr} or small institutional campuses~\cite{NYU_VPR,
		University_of_Michigan}---not the dense, mixed-use commercial districts where millions of people
	navigate daily. \textbf{(2) Daytime-only collection.}
	Most datasets capture exclusively daytime conditions.
	Illumination variation causes 50--80\% performance drops when daytime-trained models are
	deployed at night~\cite{porav2019don,anoosheh2019night}, yet balanced day--night coverage
	remains rare. \textbf{(3) Unimodal inputs.}
	Existing benchmarks rely on imagery alone, overlooking rich semantic text (store names,
	signage), video temporal dynamics, and spatial-structural relationships.
	Vision--language models such as CLIP~\cite{radford2021clip} and DINO~\cite{caron2021emerging}
	have demonstrated cross-modal benefits on curated datasets, but systematic multimodal
	evaluation in cluttered pedestrian scenes is unexplored. \textbf{(4) Short or narrow temporal spans.}
	Datasets typically cover weeks to months~\cite{sattler2018benchmarking}, preventing modeling of
	long-term urban evolution (seasonal change, commercial turnover, architectural modification).

To address these limitations, we introduce MMS-VPR for street-level VPR evaluation in dense urban environments. The dataset covers 208 fine-grained locations in Chengdu, China, containing 110,529 images and 2,527 videos aligned with semantic text annotations and graph-structured location labels. It has four key characteristics: \textbf{(1) Pedestrian-only streets}—systematic field collection in dense commercial districts inaccessible to vehicles, providing human-centric viewpoints for AR and navigation applications; \textbf{(2) Day \& night coverage}—balanced temporal sampling across daytime (7AM-5PM) and nighttime (6PM-10PM) periods, enabling illumination-robust place recognition; \textbf{(3) Multimodality}—integration of images, videos, and rich textual annotations (GPS coordinates, spatial identifiers, shop names/signage, street properties, space syntax metrics), organized in an explicit graph structure supporting diverse learning paradigms; \textbf{(4) Long temporal span}—combining systematic field collection in 2024 and seven years of public social media data (2019-2025) at the same locations, providing both fine-grained temporal granularity and extended coverage for modeling urban evolution. All released visual data are downsampled to a longest side of 256 pixels, and faces, license plates, and other personally identifiable details are anonymized before release.

Our main contributions are summarized as follows:

\begin{enumerate}[leftmargin=1.5em, itemsep=2pt, parsep=0pt, topsep=2pt]

		\item \textbf{MMS-VPR dataset}: the first multimodal street-level VPR benchmark
		systematically integrating images, videos, and semantic text with comprehensive day--night
		coverage and a 7-year temporal span in dense, pedestrian-only environments.
		The 208 fine-grained location classes---organized as an explicit spatial graph---support
		image-, video-, text-, and graph-based evaluation.
		Dataset-level annotations include GPS coordinates, OCR-extracted signage, store inventories,
		and space syntax metrics from city science (see Appendix~\ref{app:syntax}).

		\item \textbf{MMS-VPRlib}: an open-source benchmark platform providing standardized,
		reproducible pipelines for multimodal VPR across 22 baseline models spanning shallow ML,
		CNN, Transformer, GNN, and vision--language architectures.
		The platform unifies widely used VPR datasets (Pittsburgh~\cite{torii2013visual},
		Tokyo 24/7~\cite{torii201524}, Nordland~\cite{sunderhauf2013we},
		Cambridge~\cite{kendall2015posenet}, New College~\cite{smith2009newcollege}) under
		a consistent evaluation interface.

		\item \textbf{Systematic evaluation}: extensive experiments answering four research
		questions---(RQ1) benchmark performance of 22 baselines on MMS-VPR including GNN
		methods; (RQ2) modality ablation confirming multimodal gains;
		(RQ3) cross-dataset validation of MMS-VPRlib on standard benchmarks; and (RQ4)
		runtime efficiency and hyperparameter sensitivity.

	\end{enumerate}

\section{Evaluation scope and benchmark positioning}
	\label{sec:evaluation_scope}

	\subsection{Intended evaluative role}
	\label{sec:intended_role}

	\dataset{} is intended to evaluate whether a VPR system can recognize fine-grained pedestrian locations under conditions that are common in human-centered urban navigation but underrepresented in vehicle-oriented benchmarks. Formally, the main task is closed-set place recognition: given a query image, video clip, text record, or fused multimodal input, a model predicts one of $C=208$ location classes. A top-1 prediction is correct if and only if the predicted class matches the ground-truth location label. We report Accuracy, macro Precision, macro Recall, macro F1, and Recall@K.
    The benchmark supports the following evaluative claims: 1) robustness to pedestrian-scale viewpoint change and day-night illumination variation; 2) benefits and limits of multimodal inputs, especially visual media plus semantic signage or shop-name text;
3) relative behavior of generic visual backbones, VPR-specific retrieval objectives, and vision-language pretrained models; 4) reproducibility and cross-dataset comparability through a shared library.

	Notably, the benchmark does not claim to measure cross-city generalization by itself, because the physical site is a single district. It also does not provide calibrated 6DoF camera poses or an evaluation for person identification, tracking, or surveillance. Spatial graph and space-syntax annotations are released as auxiliary metadata; the main benchmark does not rank methods by those attributes unless a graph-aware protocol is explicitly selected.

	\subsection{Comparison with existing datasets and benchmark tools}
	\label{sec:positioning}

We argue that pedestrian-level VPR should be evaluated as a distinct setting rather than treated as a small variant of vehicle-view VPR. A user, robot, or AR device in a pedestrian commercial district must distinguish adjacent intersections, storefronts, alleys, and squares under four simultaneous sources of difficulty: (i) restricted human-scale viewpoints instead of vehicle panoramas; (ii) day-night illumination shifts and artificial lighting; (iii) dynamic occlusion by crowds and temporary commercial displays; and (iv) semantic ambiguity, where visually similar facades are disambiguated by text, signs, and spatial context. Existing datasets such as Pittsburgh~\cite{torii2013visual}, Tokyo 24/7~\cite{torii201524}, Nordland~\cite{sunderhauf2013we}, Cambridge~\cite{kendall2015posenet}, CV-Cities~\cite{CV_Cities}, and GSV-Cities~\cite{GSV_Cities} provide valuable evaluation resources, but they do not directly measure this dense pedestrian recognition regime.

	Table~\ref{tab:positioning} summarizes the role of \dataset{} and \library{} relative to representative VPR datasets and platforms. The key distinction is not geographic breadth alone, but the combination of pedestrian-only access, fine spatial granularity, multimodality, day-night sampling, and long-term temporal augmentation at the same locations.

	\begin{table}[t]
		\centering
		\scriptsize
		\setlength{\tabcolsep}{3pt}
		\renewcommand{\arraystretch}{1.08}
		\caption{Positioning of \dataset{} and \library{}. \cmark: supported; \pmark: partial; \xmark: not provided.}
		\label{tab:positioning}
		\resizebox{\linewidth}{!}{
			\begin{tabular}{lccccccccc}
				\toprule
				\multicolumn{10}{c}{\textbf{Panel A: Dataset comparison}}\\
				\midrule
				Dataset & Img & Vid & Txt & Day-night & Multi-view & Long-term & Pedestrian-only & Graph labels & Main setting\\
				\midrule
				Tokyo 24/7~\cite{torii201524} & \cmark & \xmark & \xmark & \cmark & \pmark & \pmark & \pmark & \xmark & city landmark\\
				Pittsburgh~\cite{torii2013visual} & \cmark & \xmark & \xmark & \xmark & \cmark & \pmark & \xmark & \xmark & street-view retrieval\\
				Nordland~\cite{sunderhauf2013we} & \cmark & \cmark & \xmark & \pmark & \xmark & \cmark & \xmark & \xmark & seasonal route\\
				Cambridge~\cite{kendall2015posenet} & \cmark & \xmark & \xmark & \cmark & \cmark & \pmark & \pmark & \xmark & landmark pose\\
				GSV-Cities~\cite{GSV_Cities} & \cmark & \xmark & \xmark & \pmark & \cmark & \cmark & \xmark & \xmark & city-scale VPR\\
				\textbf{\dataset{} (ours)} & \textbf{\cmark} & \textbf{\cmark} & \textbf{\cmark} & \textbf{\cmark} & \textbf{\cmark} & \textbf{\cmark} & \textbf{\cmark} & \textbf{\cmark} & \textbf{pedestrian fine-grained}\\
				\midrule
				\multicolumn{10}{c}{\textbf{Panel B: Benchmark platform comparison}}\\
				\midrule
				Benchmark & Img & Vid & Txt & CNN & RNN/temporal & Transformer & VLM & Multi-dataset & Unified scripts\\
				\midrule
				VPR-Bench~\cite{zaffar2021vprbench} & \cmark & \xmark & \xmark & \cmark & \cmark & \xmark & \xmark & \cmark & \cmark\\
				VPR-Ensem~\cite{joseph2025ensemble} & \cmark & \xmark & \xmark & \cmark & \cmark & \xmark & \xmark & \cmark & \cmark\\
				\textbf{\library{} (ours)} & \textbf{\cmark} & \textbf{\cmark} & \textbf{\cmark} & \textbf{\cmark} & \textbf{\cmark} & \textbf{\cmark} & \textbf{\cmark} & \textbf{\cmark} & \textbf{\cmark}\\
				\bottomrule
		\end{tabular}}
	\end{table}

	\section{Data Collection Methods}
	\label{sec:collection}

\begin{figure*}[t]
	\centering
	\includegraphics[width=\textwidth]{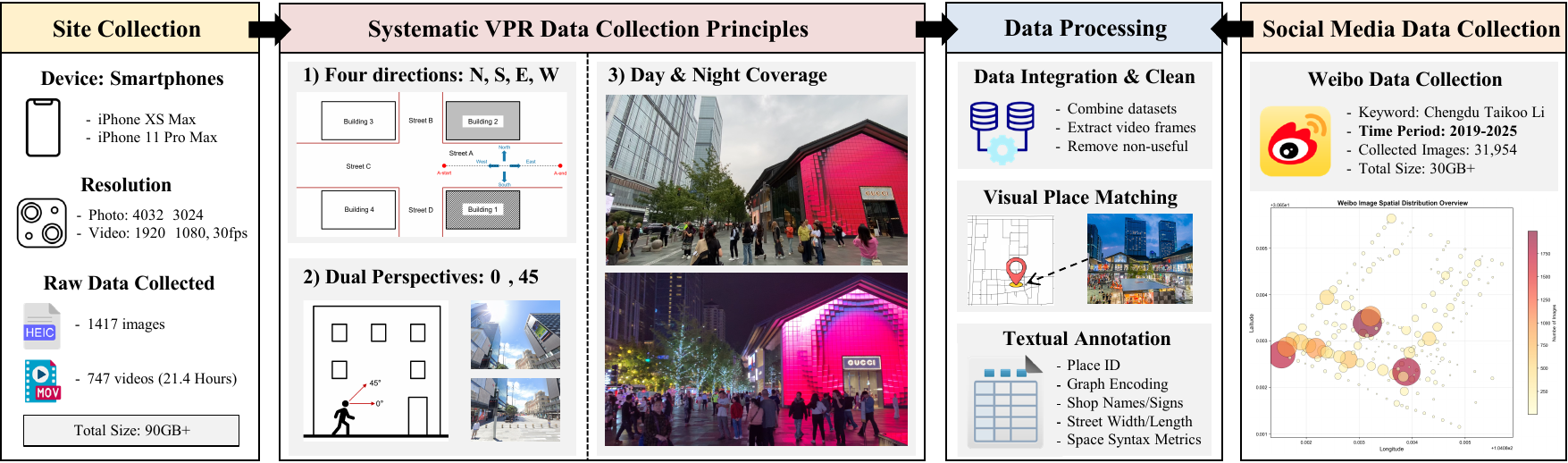}
	\caption{Pipeline for building MMS-VPR: data collection, processing, annotation and enhancement.}
	\label{fig:data_collection}
\end{figure*}

	Figure~\ref{fig:data_collection} shows the overall framework.
	The complete pipeline encompasses four stages: site collection, data processing,
	textual annotation, and social media integration.

	\subsection{Site Selection}
	\label{sec:site}

	We selected Chengdu Taikoo Li, a $\sim$70,800\,m\textsuperscript{2} open-air commercial
	district in Chengdu's CBD (Figure~\ref{fig:framework}), for three reasons:
	(1) \emph{spatial openness}---natural lighting, varied weather, and high crowd density create
	genuine environmental complexity;
	(2) \emph{functional diversity}---retail, dining, leisure, and cultural spaces ensure rich spatial
	semantics and dense pedestrian traffic, producing a high degree of visual and semantic variation over a compact area;
	(3) \emph{urban integration}---seamless connection to surrounding urban pulic streets makes the environment representative of globally common
	pedestrianized commercial districts.
	Crucially, this environment challenges place recognition with \emph{repetitive visual elements}
	(franchise storefronts, identical signage across locations) and dense occlusion patterns
	that indoor mall datasets~\cite{lee2021large} and large-area outdoor datasets both fail to capture.
	Detailed site characteristics are in Appendix~\ref{app:site_details}.

	\subsection{Systematic Data Collection Framework}

	To maximize reproducibility, we use consumer smartphones (iPhone XS Max, 11 Pro Max)
	for all visual capture: 4032$\times$3024 images and 1920$\times$1080 video at 30fps.
	The framework is grounded in three evidence-based principles
	(Table~\ref{tab:collection-comparison}).

	\noindent\textbf{Principle 1: Four-Direction Coverage.}
	VPR performance degrades up to 40\% under viewpoint mismatch~\cite{berton2021viewpoint}.
	We systematically capture all four cardinal directions (N, S, E, W) for each street segment,
	providing coverage matching real-world social media content patterns rather than
	360$^\circ$ panoramas that introduce geometric distortion~\cite{torii201524}.

	\noindent\textbf{Principle 2: Dual-Perspective Capture.}
	Pedestrians naturally employ two viewing strategies in high-rise environments:
	forward-looking (0$^\circ$) for navigation and upward-looking (30--60$^\circ$) for landmark
	recognition~\cite{palmer1999vision,cutting2003reconceiving}.
	Incorporating upward-facing imagery improves place recall by 15--30\% where building
	facades serve as primary discriminative features~\cite{hauagge2012image3d}.
	We capture each view at horizontal (0$^\circ$) and upward (45$^\circ$) angles.

	\noindent\textbf{Principle 3: Balanced Day--Night Coverage.}
	Illumination variation causes 50--80\% performance drops for daytime-trained
	models~\cite{porav2019don,anoosheh2019night}.
	Social-media imagery in commercial districts exhibits pronounced temporal patterns tied to
	visitor activity rhythms~\cite{vu2015exploring}.
	We enforce equivalent data volume during daytime (7AM--5PM) and nighttime (6PM--10PM).

\begin{table}[t]
		\centering
		\caption{Data collection strategy comparison.}
		\label{tab:collection-comparison}
		\scriptsize
		\setlength{\tabcolsep}{3pt}
		\renewcommand{\arraystretch}{1.1}
		\begin{tabular}{@{}lcccc@{}}
    \toprule
    \textbf{Dataset} & \textbf{Viewpoint} & \textbf{Perspective}
    & \textbf{Day/Night} & \textbf{Rationale} \\
    \midrule
    Google Street View~\cite{anguelov2010google} & 360° & Single (elevated) & Limited & Road network vehicle coverage \\
    Mapillary~\cite{Mapillary_Vistas} & Opportunistic & Variable & Opportunistic & Crowdsourced urban imagery \\
    RobotCar~\cite{maddern20171} & Forward-facing & Single (vehicle) & Systematic & Autonomous driving research \\
    Pittsburgh~\cite{torii2013visual} & Panoramic & Single & Daytime only & Urban street-view mining \\
    Nordland~\cite{sunderhauf2013we} & Forward-facing & Single (rail) & Seasonal & Seasonal railway traversal \\
    \hdashline
    \textbf{MMS-VPR (ours)} & \textbf{4 directions} & \textbf{Dual (0°, 45°)}
    & \textbf{Balanced}
    & \textbf{Human vision~\cite{palmer1999vision}; VPR studies~\cite{lowry2015visual}; Social media~\cite{hays2008im2gps}} \\
    \bottomrule
\end{tabular}
	\end{table}

	\subsection{Dual-Source Data Integration}

	\noindent\textbf{Field Collection (2024).}
	Systematic on-site collection yielded 78,575 images and 2,527 video clips
	across all 208 locations, with fine-grained temporal coverage from 7AM--10PM and varied
	pedestrian occlusion patterns.
	Raw data totaled $>$90\,GB; frames extracted at 1\,Hz from 747 raw videos.

	\noindent\textbf{Social Media Integration (2019--2025).}
	To address limited temporal span, we curated 31,954 georeferenced images
	from Weibo (Chinese Twitter) matching keyword ``Chengdu Taikoo Li'' and localized
	to our 208 locations, spanning January 2019 through December 2025.
	This enables modeling of seasonal variation, architectural change, and commercial
	turnover over seven years.
	Social media imagery has been widely adopted to augment VPR with temporal and
	viewpoint diversity~\cite{City_Scale,hays2008im2gps}.

	The integrated dataset contains \textbf{110,529 images} and \textbf{2,527 videos}
	across \textbf{208 location classes}.

    All field data were collected with \textit{official permission from Chengdu Taikoo Li}
	for non-commercial academic research and dataset publication.
	Collection was strictly limited to publicly accessible outdoor pedestrian areas.
	All images---field-collected and Weibo-sourced---underwent strict face detection and
	pixelation, license plate masking and downsampling (maximum edge of 256\,pixels).

	\section{MMS-VPR Dataset}
	\label{sec:dataset}

	\subsection{Task Formulation}
	\label{sec:task}

	We formulate VPR as closed-set classification over $C=208$ locations.
	Each location constitutes one class, uniquely identified by its position in the spatial graph.
	Given a query (image, video, or text, or any combination), a model predicts class
	$\hat{c} \in \{0, \ldots, 207\}$.
	A prediction is correct if and only if $\hat{c}$ matches the ground-truth location label.
	Performance is measured by Accuracy, Precision, Recall, and F1 over a held-out test split
	(per-location 80/20 stratified split ensuring each class appears in both training and test).
	This closed-set formulation reflects the intended deployment scenario: a system trained
	on a known urban district that must perform real-time pedestrian localization within it.

	\subsection{Multimodal Data Composition}

	MMS-VPR integrates three complementary modalities unified through a spatial graph:

	\noindent\textbf{Image Data.}
	110,529 georeferenced images from two sources:
	(1) \textit{Field-collected}---78,575 images including raw photographs and 1\,Hz-extracted frames,
	covering balanced day--night periods under four-direction, dual-perspective protocols;
	(2) \textit{Social media}---31,954 curated photographs from Weibo (2019--2025).
	Resolution is down-scaled from 4032$\times$3024 (original) to 256$\times$192 (preprocessed).

	\noindent\textbf{Video Data.}
	2,527 video clips (20--60 seconds, 256$\times$144 at 30fps).
	Videos provide motion dynamics absent in stills, supporting temporal architectures
	(LSTM, GRU, 3D-CNN, Video Transformers).

	\noindent\textbf{Textual Data.}
	Each location is annotated with: (1) \textit{Spatial identifiers} encoding graph topology
	(e.g., ``Eh-5-3''), enabling adjacency matrix construction for GNNs;
	(2) \textit{Visible signage}---discriminative store names and landmark text from
	OCR and manual verification (e.g., ``Starbucks, Adidas, LEGO'');
	(3) \textit{GPS coordinates};
	(4) \textit{Space syntax metrics}---quantitative spatial configuration from city science
	(provided for future research; see Appendix~\ref{app:syntax}).

	\subsection{Graph-based Spatial Structure}
All 208 locations form a spatial graph $G=(V,E)$ representing the pedestrian network topology,
	comprising four spatial types (Table~\ref{tab:stats}):
	\textbf{Nodes} ($|V_{node}|=81$)—street intersections encoded as ``N-$i$-$j$'';
	\textbf{Horizontal Edges} ($|V_h|=61$)—east--west street segments encoded as ``Eh-$i$-$j$'';
	\textbf{Vertical Edges} ($|V_v|=64$)—north--south street segments encoded as ``Ev-$j$-$i$'';
	\textbf{Squares} ($|V_s|=2$)—open-area sub-graphs encoded as ``S-$k$''. This hierarchical encoding preserves both geometric position and topological relationships,
	supporting graph-based learning.
	We provide complete connectivity tables (edge connections, shared nodes, turning angles, distances)
	enabling direct adjacency matrix construction for GNN-based methods.
	Detailed encoding methodology is in Appendix~\ref{app:encoding_methodology}.

\begin{table}[t]
    \caption{Dataset statistics by spatial type.}
    \label{tab:stats}
    \centering
    \small
    \setlength{\tabcolsep}{6pt}
    \renewcommand{\arraystretch}{1.05}
    \begin{tabular}{@{}lccccc@{}}
        \toprule
        Spatial Type & Index & Classes & Images & Videos & Average per Location \\
        \midrule
        Nodes (street intersections)         & 0--80    & 81  & 47,049  & 1,197 & 580.8   \\
        Squares (large open spaces)          & 81--82   & 2   & 5,686   & 7     & 2,843.0 \\
        Horizontal Edges (east-west streets) & 83--143  & 61  & 28,178  & 662   & 461.9   \\
        Vertical Edges (north-south streets) & 144--207 & 64  & 29,616  & 661   & 462.7   \\
        \textbf{Total} & \textbf{0--207} & \textbf{208} & \textbf{110,529} & \textbf{2,527} & \textbf{531.4} \\
        \bottomrule
    \end{tabular}
\end{table}

	\section{MMS-VPRlib Benchmark}
	\label{sec:lib}

\begin{figure*}[t]
		\centering
		\includegraphics[width=\linewidth]{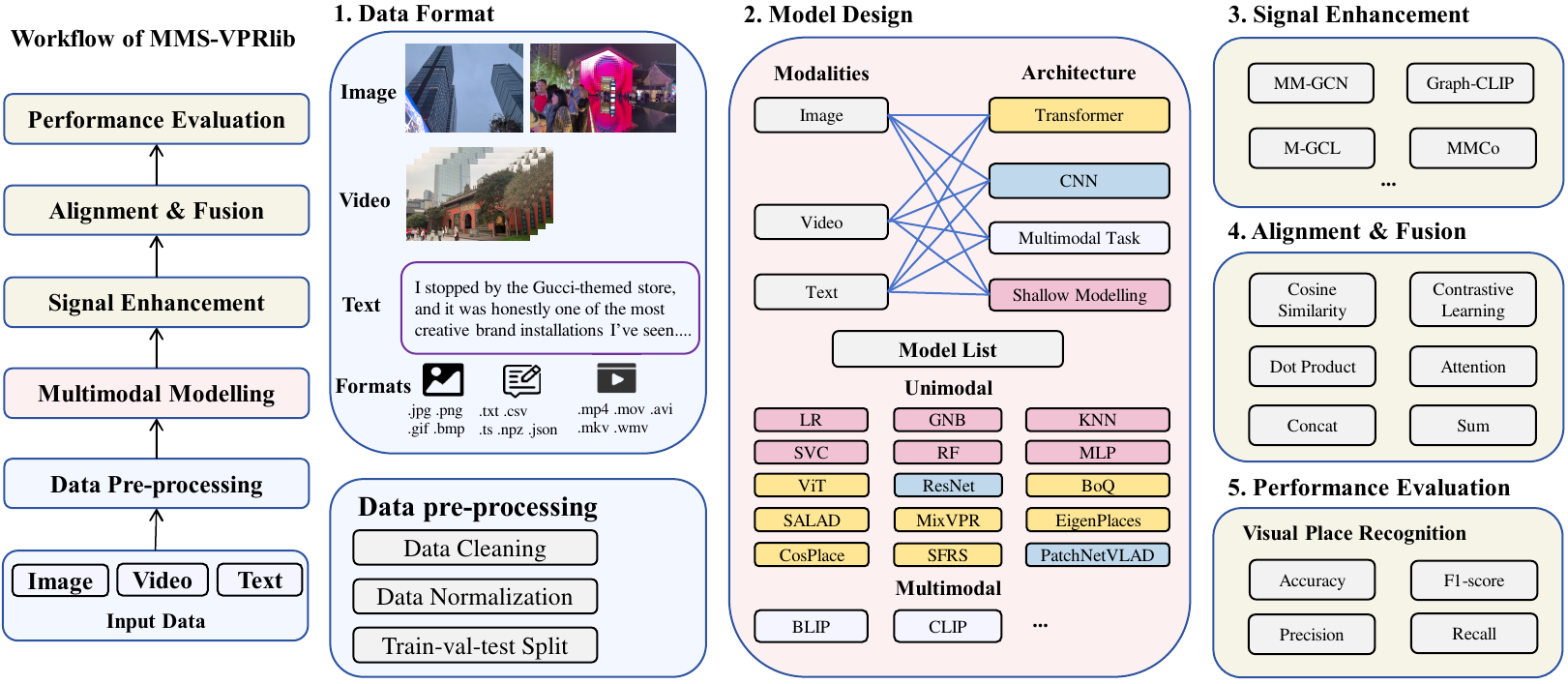}
		\caption{Workflow of MMS-VPRlib: modular pipeline for multimodal, graph-aware VPR evaluation.}
		\label{fig:workflow_vpr}
	\end{figure*}

In view of the limited support for modern algorithms and datasets in existing VPR benchmarks (Table~\ref{tab:positioning}, Panel B), we propose MMS-VPRlib, a unified benchmark supporting multimodal inputs, diverse VPR datasets, and state-of-the-art VPR models. Unlike VPR-Bench~\cite{zaffar2021vprbench} and VPR-Ensem~\cite{joseph2025ensemble}, MMS-VPRlib provides a comprehensive and extensible framework for diverse VPR tasks (Figure~\ref{fig:workflow_vpr}).

	\textbf{Data Format and Pre-processing.}
	MMS-VPRlib accepts PNG, JPEG, MP4, and TXT inputs.
	Pre-processing includes resolution standardization, 1\,Hz frame extraction for video,
	modality-specific normalization, and principled 80/20 per-location train--test splitting.
	Multi-modal data undergo signal enhancement (low-light correction, deblurring,
	text normalization) to reduce spurious dataset bias.

	\textbf{Model Design.}
	We provide four model families:
	(i) \emph{Shallow ML}: LR, GNB, SVC, KNN, RF, MLP operating on PCA-reduced features;
	(ii) \emph{CNN-based}: ResNet~\cite{resnet}, PatchNetVLAD~\cite{patchnetvlad};
	(iii) \emph{Transformer-based}: ViT~\cite{vit}, R2Former~\cite{zhu2023r2former}, SALAD~\cite{SALAD},
	BoQ~\cite{boq}, CosPlace~\cite{CosPlace}, MixVPR~\cite{mixvpr},
	EigenPlaces~\cite{EigenPlaces};
	(iv) \emph{Graph neural networks}: GCN~\cite{GCN}, GAT~\cite{GAT}, HGNN~\cite{hgnn},
	HGNN+ResNet (multimodal fusion);
	(v) \emph{Vision--language models}: CLIP~\cite{radford2021clip}, BLIP~\cite{li2022blip}.
	Text is encoded via pretrained BERT~\cite{devlin2019bert} (outperforming GPT-2 variants
	in preliminary experiments).

	\textbf{Alignment, Fusion, and Evaluation.}
	Cross-modal inputs are aligned into a shared embedding space.
	Fusion supports both feature concatenation and score-level aggregation.
	Evaluation is standardized to Accuracy, Precision, Recall, and F1,
	with mean and std over 5 runs.

	\textbf{Supported Datasets.}
	MMS-VPRlib supports multiple existing datasets: Tokyo 24/7~\cite{torii201524},
	Pittsburgh~\cite{torii2013visual}, Cambridge~\cite{kendall2015posenet},
	New College~\cite{smith2009newcollege}, Nordland~\cite{sunderhauf2013we},
	CityPlace~\cite{cummins2008fabmap}, and our MMS-VPR.

	\section{Experiments}
	\label{sec:exp}

	\noindent\textbf{Experimental scope.}
	The experiments below evaluate MMS-VPR as a \textit{benchmark}---assessing how well existing
	methods perform and where they fall short---and validate MMS-VPRlib as a unified evaluation
	platform.

	\subsection{Research Questions}

	\begin{itemize}[leftmargin=1.5em, itemsep=1pt, topsep=3pt]
		\item \textbf{RQ1}: How do existing VPR-related baselines perform on MMS-VPR data
		set?
		\item \textbf{RQ2}: How effective are individual modalities and their fusion in improving MMS-VPR performance?
		\item \textbf{RQ3}: Can MMS-VPRlib reproduce reasonable results on standard VPR datasets?
		\item \textbf{RQ4}: How robust is the proposed framework with respect to computational efficiency and hyperparameter variations?
	\end{itemize}

	\subsection{Setup}

\textbf{22 Baseline Models:} \textit{1) Shallow ML:} LR~\cite{LR}, GNB~\cite{GNB}, SVC~\cite{SVC}, KNN~\cite{KNN},
RF~\cite{RF}, MLP~\cite{MLP}; \textit{2) CNN-based:} ResNet~\cite{resnet}, PatchNetVLAD~\cite{patchnetvlad}; \textit{3) Transformer-based:} ViT~\cite{vit}, R2Former~\cite{zhu2023r2former}, BoQ~\cite{boq},
MixVPR~\cite{mixvpr}, SALAD~\cite{SALAD}, SFRS~\cite{ge2020self},
EigenPlaces~\cite{EigenPlaces}, CosPlace~\cite{CosPlace}; \textit{4) GNN-based:} GCN~\cite{GCN}, GAT~\cite{GAT}, HGNN~\cite{hgnn},
		HGNN+ResNet; \textit{5) Vision--Language:} CLIP~\cite{radford2021clip}, BLIP~\cite{li2022blip}.

\textbf{Hyperparameters:}
The major hyperparameter settings of representative baselines are summarised in Table~\ref{tab:hyperparams_remaining} from Appendix~\ref{app:hyper}. Specifically, ViT~\cite{vit} adopts 6 Transformer layers with an embedding dimension of 64 and 8 attention heads, ResNet~\cite{resnet} uses a 50-layer backbone with a 2048-dimensional feature output, and CosPlace~\cite{CosPlace} employs GeM pooling with a 512-dimensional descriptor and a CosFace margin of 0.35. In addition, PatchNetVLAD~\cite{patchnetvlad} uses 16 VLAD clusters with multi-scale grid aggregation, and MixVPR~\cite{mixvpr} adopts 3 Mix blocks with a 256-dimensional projection space.

\subsection{RQ1: Benchmark Results on MMS-VPR}
\label{sec:rq1}

\begin{table*}[t]
    \caption{
    \textbf{22 Baseline results on MMS-VPR} (full 208 classes, image-only input,
    80/20 split, mean over 5 runs).
    \textbf{Bold}: best per metric; \textit{italic}: second best.
    $^\dagger$Pure GNNs use graph-structured inputs (no visual features),
    serving as structural-only baselines; HGNN+ResNet denotes graph-visual fusion.}
    \label{tab:baseline}
    \centering
    \scriptsize
    \setlength{\tabcolsep}{3.5pt}
    \renewcommand{\arraystretch}{1.1}
    \resizebox{\textwidth}{!}{
    \begin{tabular}{l *{6}{c} *{4}{c} *{2}{c} *{8}{c} *{2}{c}}
        \toprule
        & \multicolumn{6}{c}{\textit{Shallow ML}}
        & \multicolumn{4}{c}{\textit{GNN}}
        & \multicolumn{2}{c}{\textit{CNN}}
        & \multicolumn{8}{c}{\textit{Transformer / VPR}}
        & \multicolumn{2}{c}{\textit{Vision--Lang}} \\
        \cmidrule(lr){2-7}\cmidrule(lr){8-11}\cmidrule(lr){12-13}
        \cmidrule(lr){14-21}\cmidrule(lr){22-23}
        Metric
        & LR & GNB & SVC & KNN & RF & MLP
        & GCN$^\dagger$ & GAT$^\dagger$ & HGNN$^\dagger$ & H+R
        & ResNet & PNet
        & ViT & R2F & BoQ & SFRS & MixVPR & SALAD & \textit{EigPl} & \textbf{CosPl}
        & BLIP & CLIP \\
        \midrule
        Acc
        & 0.110 & 0.082 & 0.080 & 0.422 & 0.198 & 0.202
        & 0.076 & 0.078 & 0.078 & 0.879
        & 0.856 & 0.846
        & 0.596 & 0.771 & 0.744 & 0.734 & 0.789 & 0.870 & \textit{0.918} & \textbf{0.933}
        & 0.674 & 0.885 \\
        P
        & 0.090 & 0.090 & 0.040 & 0.508 & 0.396 & 0.184
        & 0.048 & 0.046 & 0.024 & 0.879
        & 0.849 & 0.838
        & 0.584 & 0.775 & 0.720 & 0.722 & 0.793 & 0.870 & \textit{0.910} & \textbf{0.925}
        & 0.678 & 0.886 \\
        R
        & 0.060 & 0.100 & 0.030 & 0.402 & 0.154 & 0.184
        & 0.034 & 0.033 & 0.032 & 0.872
        & 0.837 & 0.860
        & 0.564 & 0.770 & 0.694 & 0.673 & 0.785 & 0.865 & \textit{0.904} & \textbf{0.921}
        & 0.683 & 0.878 \\
        F1
        & 0.054 & 0.084 & 0.020 & 0.406 & 0.190 & 0.174
        & 0.028 & 0.024 & 0.020 & 0.874
        & 0.841 & 0.845
        & 0.566 & 0.765 & 0.703 & 0.689 & 0.781 & 0.861 & \textit{0.906} & \textbf{0.924}
        & 0.671 & 0.877 \\
        \bottomrule
    \end{tabular}}
    \begin{flushleft}
        \scriptsize
        \textit{Abbreviations}: PNet\,=\,PatchNetVLAD~\cite{patchnetvlad}; R2F\,=\,R2Former~\cite{zhu2023r2former};
    EigPl\,=\,EigenPlaces~\cite{EigenPlaces}; CosPl\,=\,CosPlace~\cite{CosPlace}; H+R\,=\,HGNN+ResNet.
    \end{flushleft}
\end{table*}

Table~\ref{tab:baseline} presents results across all 22 baselines.
\textbf{VPR-specialized CosPlace achieves the strongest result} (Acc: 0.933, F1: 0.924),
outperforming the strongest generic backbone ResNet by 9.0\,percentage points accuracy (0.856$\to$0.933),
confirming VPR-oriented metric learning and aggregation provides additional gains beyond generic visual feature extraction.
Among VPR-specialized methods, EigenPlaces ranks second (Acc: 0.918), followed closely
by SALAD (Acc: 0.870) and SFRS (Acc: 0.734).
Large-scale pretraining substantially helps: \textbf{CLIP improves vanilla ViT by 48.5\,\%}
(0.596$\to$0.885).
Pure GNN methods (GCN, GAT, HGNN) receive only structural inputs without visual features,
serving as structural-only lower bounds baseline.
Critically, \textbf{HGNN+ResNet (Acc: 0.879) outperforms a standalone ResNet (Acc: 0.856)},
confirming that graph-structured spatial information provides complementary signal to visual features in the appropriate evaluation setting.

\subsection{RQ2: Modality Ablation}
\label{sec:rq2}

To isolate the contribution of each modality independently of model capacity, we adopt a
lightweight version of \textit{ResNet}~\cite{resnet} , evaluated on $C=208$ classes of the field-collected dataset
across all modalities (image, video, text and embedded graph structure).
Image and video modalities are extracted from this frozen encoder, while text feature is
obtained from a frozen pretrained BERT~\cite{devlin2019bert} encoder. Fusion weights
between modalities are optimized jointly with the classifier head.

\begin{figure}[t]
    \centering
    \begin{subfigure}[t]{0.39\columnwidth}
        \centering
        \includegraphics[width=\linewidth]{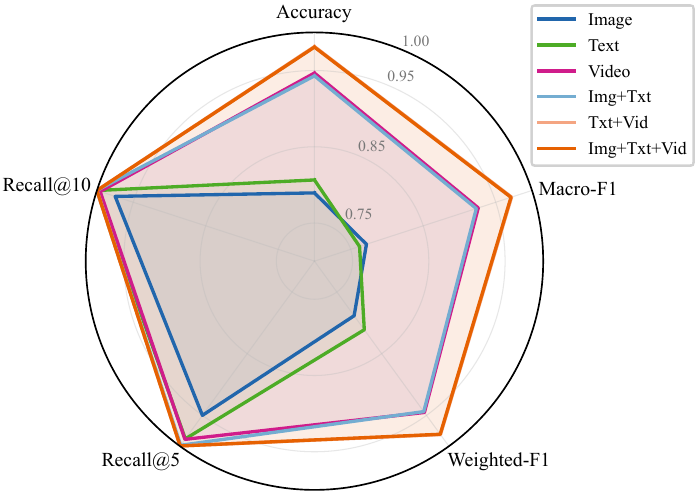}
        \caption{Radar chart of key metrics of modalities.}
        \label{fig:radar_ablation}
    \end{subfigure}\hfill
    \begin{subfigure}[t]{0.59\columnwidth}
        \centering
        \includegraphics[width=\linewidth]{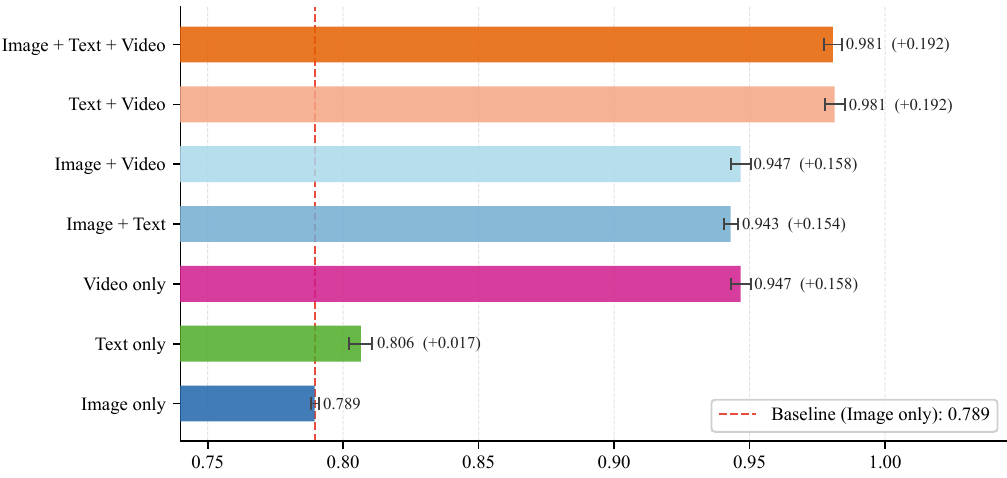}
        \caption{Accuracy gain over image-only baseline
                 (mean\,$\pm$\,std, 3 splits).}
        \label{fig:incremental_gain}
    \end{subfigure}
    \caption{Multimodal ablation experiment result visualizations on MMS-VPR dataset.}
    \label{fig:ablation_visual}
\end{figure}

Figure~\ref{fig:radar_ablation}, and Figure~\ref{fig:incremental_gain} together show the modality ablation results.
Overall, the ablation study demonstrate that each modality is indispensable for improving MMS-VPR performance. Compared with the Image-only setting (0.789), incorporating video increases the accuracy to 0.947, corresponding to an improvement of 20.0\%, while further integrating Text raises the accuracy to 0.981, yielding an additional improvement of 3.6\%. Among all modalities, Video contributes the most substantial performance gain relative to static images, highlighting the importance of temporal continuity and motion dynamics for fine-grained place recognition.

\subsection{RQ3: Cross-Dataset Validation of MMS-VPRlib}

	\begin{table*}[t]
		\centering
		\scriptsize
		\setlength{\tabcolsep}{3pt}
		\caption{\textbf{Cross-dataset results} on five VPR datasets.
			Best/second-best highlighted in \textbf{bold}/\underline{underline}.}
		\label{tab:unimodal}
		\resizebox{\textwidth}{!}{
			\begin{tabular}{l*{20}{c}}
				\toprule
				\multirow{2}{*}{Model}
				& \multicolumn{4}{c}{Tokyo~\cite{torii201524}} & \multicolumn{4}{c}{New College~\cite{smith2009newcollege}}
				& \multicolumn{4}{c}{Pittsburgh~\cite{torii2013visual}} & \multicolumn{4}{c}{Nordland~\cite{sunderhauf2013we}}
				& \multicolumn{4}{c}{Cambridge~\cite{kendall2015posenet}} \\
				\cmidrule(lr){2-5}\cmidrule(lr){6-9}\cmidrule(lr){10-13}
				\cmidrule(lr){14-17}\cmidrule(lr){18-21}
				& Acc & P & R & F1 & Acc & P & R & F1 & Acc & P & R & F1
				& Acc & P & R & F1 & Acc & P & R & F1 \\
				\midrule
				PatchNetVLAD~\cite{patchnetvlad}
				& 0.846 & 0.838 & 0.860 & 0.845
				& 0.144 & 0.009 & 0.038 & 0.013
				& 0.870 & 0.860 & 0.887 & 0.860
				& 0.430 & 0.420 & 0.449 & 0.410
				& 0.569 & 0.207 & 0.321 & 0.250 \\
				R2Former~\cite{zhu2023r2former}
				& 0.771 & 0.775 & 0.770 & 0.765
				& 0.722 & 0.453 & 0.429 & 0.428
				& 0.910 & 0.910 & 0.911 & 0.910
				& 0.710 & 0.710 & 0.720 & 0.710
				& 0.946 & 0.943 & 0.923 & 0.932 \\
				ViT~\cite{vit}
				& 0.606 & 0.602 & 0.612 & 0.599
				& 0.641 & 0.447 & 0.484 & 0.452
				& 0.701 & 0.698 & 0.712 & 0.701
				& 0.523 & 0.521 & 0.531 & 0.522
				& 0.815 & 0.764 & 0.752 & 0.756 \\
				BLIP~\cite{li2022blip}
				& 0.674 & 0.678 & 0.683 & 0.671
				& 0.686 & 0.527 & 0.477 & 0.480
				& 0.761 & 0.772 & 0.782 & 0.771
				& 0.583 & 0.592 & 0.603 & 0.592
				& 0.818 & 0.793 & 0.714 & 0.739 \\
				MixVPR~\cite{mixvpr}
				& 0.789 & 0.793 & 0.785 & 0.781
				& 0.758 & 0.549 & 0.533 & 0.530
				& 0.904 & 0.912 & 0.915 & 0.917
				& 0.730 & 0.730 & 0.762 & 0.730
				& 0.971 & 0.971 & 0.959 & 0.965 \\
				EigenPlaces~\cite{EigenPlaces}
				& \underline{0.927} & \underline{0.929} & \underline{0.930} & \underline{0.921}
				& 0.773 & 0.587 & 0.583 & 0.571
				& \textbf{0.920} & \textbf{0.920} & \textbf{0.925} & \textbf{0.920}
				& 0.681 & 0.689 & 0.712 & 0.694
				& 0.968 & 0.954 & 0.951 & 0.953 \\
				ResNet~\cite{resnet}
				& 0.794 & 0.798 & 0.801 & 0.790
				& 0.775 & \underline{0.633} & \underline{0.618} & \underline{0.611}
				& 0.879 & 0.881 & 0.889 & 0.879
				& \textbf{0.861} & \textbf{0.830} & \underline{0.814} & \underline{0.815}
				& 0.968 & 0.963 & 0.958 & 0.959 \\
				CosPlace~\cite{CosPlace}
				& 0.816 & 0.819 & 0.822 & 0.812
				& 0.781 & 0.588 & 0.596 & 0.580
				& 0.900 & 0.900 & 0.909 & 0.900
				& 0.560 & 0.560 & 0.585 & 0.560
				& 0.974 & \underline{0.977} & 0.953 & 0.964 \\
				CLIP~\cite{radford2021clip}
				& 0.843 & 0.846 & 0.851 & 0.839
				& \underline{0.790} & 0.500 & 0.529 & 0.503
				& 0.901 & 0.899 & 0.903 & 0.897
				& 0.742 & 0.741 & 0.753 & 0.742
				& 0.930 & 0.931 & 0.914 & 0.917 \\
				BoQ~\cite{boq}
				& \textbf{0.975} & \textbf{0.977} & \textbf{0.981} & \textbf{0.973}
				& 0.787 & 0.494 & 0.533 & 0.498
				& \underline{0.920} & \underline{0.920} & \underline{0.924} & \underline{0.920}
				& \underline{0.817} & \underline{0.821} & \textbf{0.831} & \textbf{0.826}
				& \underline{0.984} & 0.976 & \underline{0.973} & \underline{0.974} \\
				SALAD~\cite{SALAD}
				& 0.804 & 0.807 & 0.800 & 0.795
				& \textbf{0.844} & \textbf{0.711} & \textbf{0.724} & \textbf{0.711}
				& 0.611 & 0.644 & 0.621 & 0.627
				& 0.744 & 0.727 & 0.763 & 0.709
				& \textbf{0.987} & \textbf{0.985} & \textbf{0.987} & \textbf{0.986} \\
				\bottomrule
		\end{tabular}}
	\end{table*}

	MMS-VPRlib successfully reproduces state-of-the-art results on five standard benchmarks
	(Table~\ref{tab:unimodal}).
BoQ leads on Tokyo~\cite{torii201524} (0.975) and Nordland~\cite{sunderhauf2013we} (0.817);SALAD leads on Cambridge~\cite{kendall2015posenet} (0.987)
	and New College~\cite{smith2009newcollege} (0.844).
	These figures are consistent with published results~\cite{boq,SALAD}, confirming that
	MMS-VPRlib provides reliable, fair evaluation. Comparing to MMS-VPR: CosPlace achieves \textit{0.974} on Cambridge~\cite{kendall2015posenet} but only \textbf{0.933} on MMS-VPR;
	SALAD achieves \textit{0.987} on Cambridge~\cite{kendall2015posenet} but only \textbf{0.870} on MMS-VPR.
	This performance inversion highlights that fine-grained
	pedestrian recognition poses distinct challenges from large-area benchmarks and requires dedicated evaluation.

	\subsection{RQ4: Runtime, Memory, and Sensitivity}

	\begin{figure}[t]
		\centering
		\begin{subfigure}[t]{0.52\columnwidth}
			\centering
			\includegraphics[width=\linewidth]{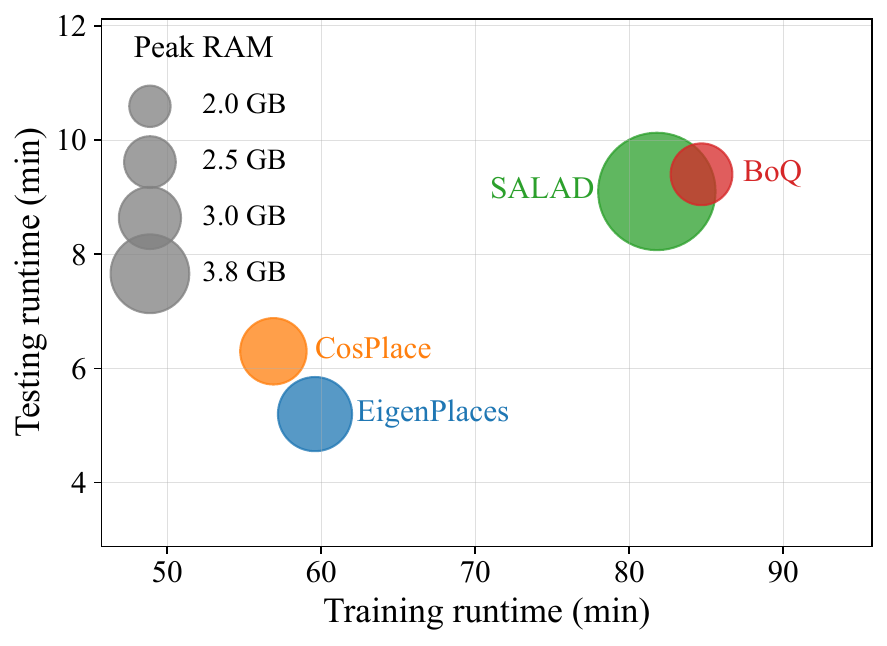}
			\caption{Runtime--memory (bubble $\propto$ RAM).}
			\label{fig:time_ram}
		\end{subfigure}\hfill
		\begin{subfigure}[t]{0.46\columnwidth}
			\centering
			\includegraphics[width=\linewidth]{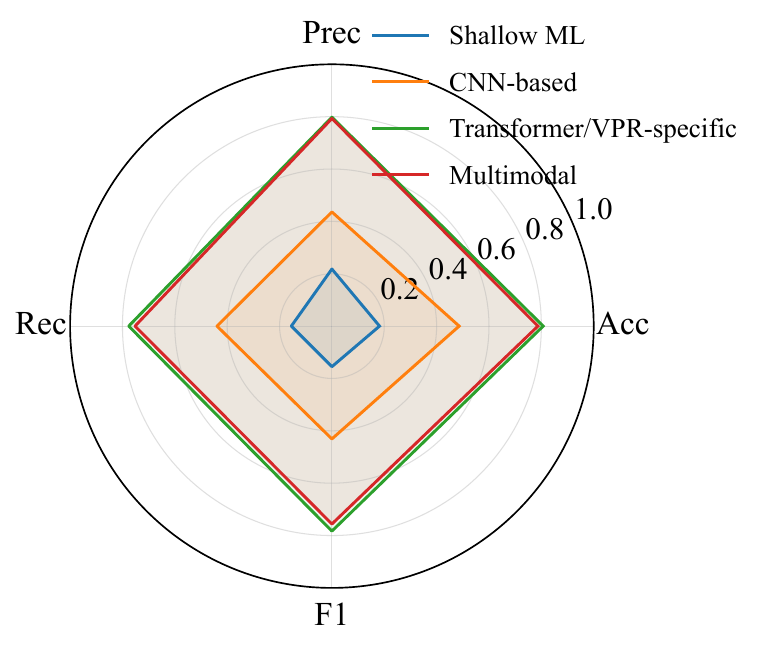}
			\caption{Category-mean performance radar.}
			\label{fig:radar}
		\end{subfigure}
		\caption{Efficiency profiles on MMS-VPR.}
		\label{fig:efficiency}
	\end{figure}

\begin{figure*}[t]
		\centering
		\begin{subfigure}[t]{0.24\textwidth}
			\includegraphics[width=\linewidth]{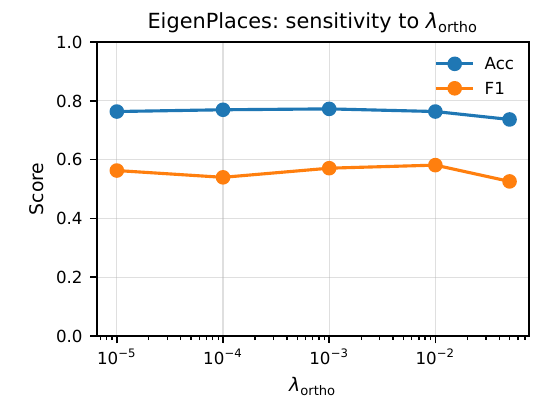}
			\caption{EigenPlaces: $\lambda_{\mathrm{ortho}}$}
		\end{subfigure}\hfill
		\begin{subfigure}[t]{0.24\textwidth}
			\includegraphics[width=\linewidth]{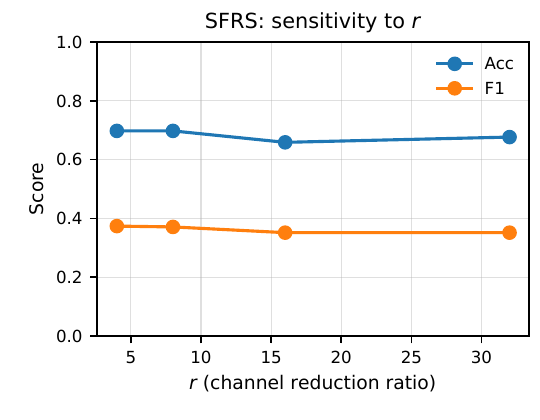}
			\caption{SFRS: $r$}
		\end{subfigure}\hfill
		\begin{subfigure}[t]{0.24\textwidth}
			\includegraphics[width=\linewidth]{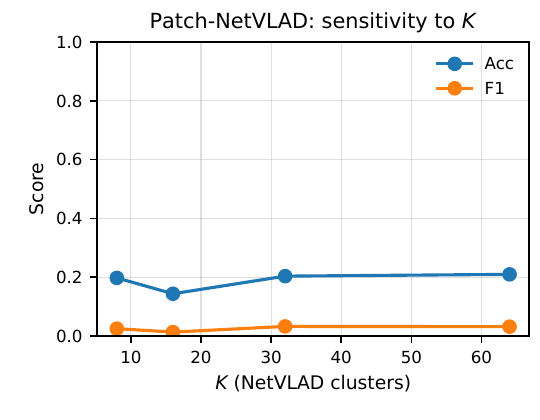}
			\caption{PatchNetVLAD: $K$}
		\end{subfigure}\hfill
		\begin{subfigure}[t]{0.24\textwidth}
			\includegraphics[width=\linewidth]{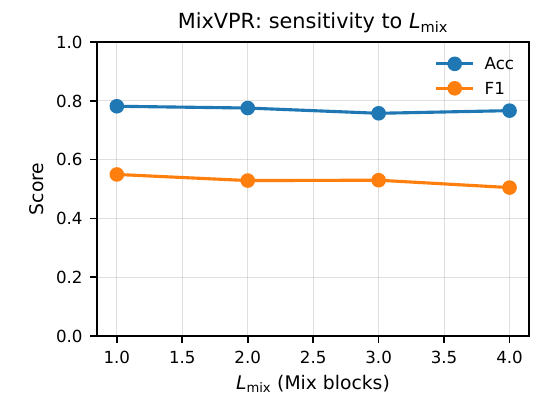}
			\caption{MixVPR: $L_{\mathrm{mix}}$}
		\end{subfigure}

		\vspace{2mm}

		\begin{subfigure}[t]{0.24\textwidth}
			\includegraphics[width=\linewidth]{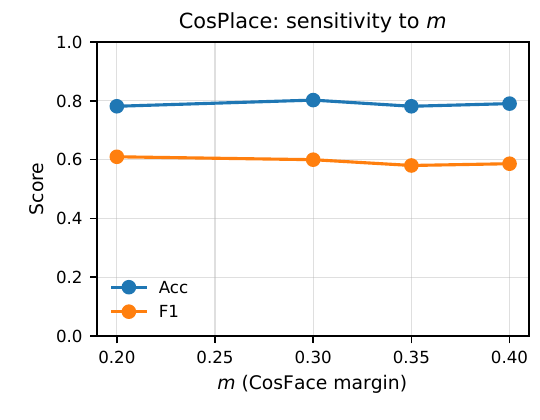}
			\caption{CosPlace: $m$}
		\end{subfigure}\hfill
		\begin{subfigure}[t]{0.24\textwidth}
			\includegraphics[width=\linewidth]{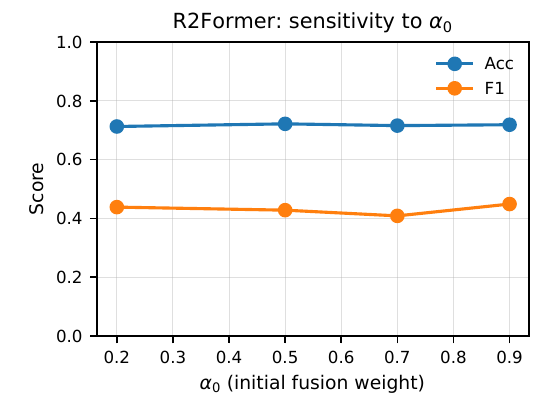}
			\caption{R2Former: $\alpha_0$}
		\end{subfigure}\hfill
		\begin{subfigure}[t]{0.24\textwidth}
			\includegraphics[width=\linewidth]{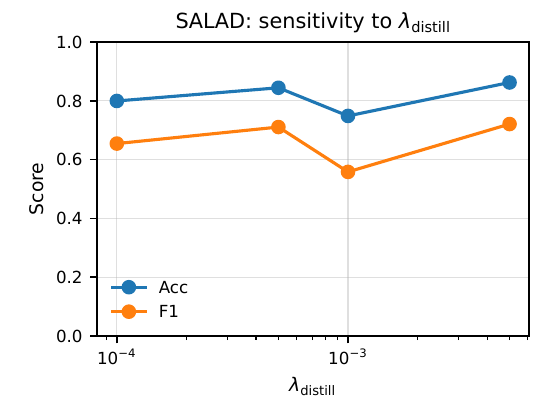}
			\caption{SALAD: $\lambda_{\mathrm{distill}}$}
		\end{subfigure}\hfill
		\begin{subfigure}[t]{0.24\textwidth}
			\includegraphics[width=\linewidth]{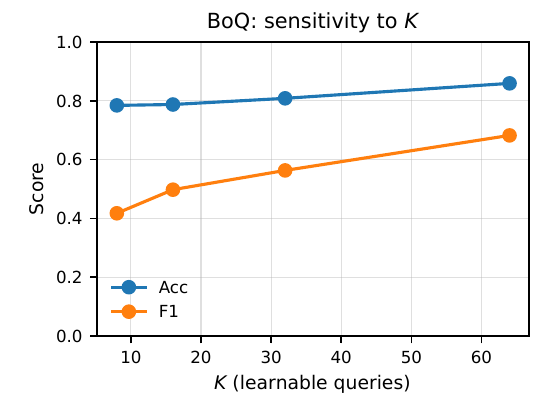}
			\caption{BoQ: $K$}
		\end{subfigure}
		\caption{Sensitivity analysis: Acc and F1 vs.\ key hyperparameters across 8 representative models.}
		\label{fig:sensitivity}
	\end{figure*}

	Figure~\ref{fig:efficiency} illustrates the trade-offs between computational efficiency and performance. CosPlace~\cite{CosPlace} and EigenPlaces~\cite{EigenPlaces} achieve high accuracy with relatively low runtime and memory consumption ($<$3.5,GB), making them practical choices under hardware constraints, whereas SALAD~\cite{SALAD} incurs the highest memory cost but delivers the best performance on sequential datasets. Overall, Transformer-based and VPR-specific architectures consistently dominate the radar plot across most evaluation metrics, while shallow machine learning and purely GNN-based approaches lag substantially behind. The sensitivity analysis in Figure~\ref{fig:sensitivity} further demonstrates stable operating regions across a broad range of hyperparameter settings: BoQ is most sensitive to the query count, achieving optimal performance at $K=64$, whereas CosPlace performs best with smaller margin values ($m \approx 0.2$--0.3$).$ Comprehensive runtime statistics are provided in Appendix~\ref{app:runtime}.

	\section{Conclusion}
	\label{sec:conclusion}

	We present MMS-VPR, a fine-grained multimodal benchmark that elevates VPR difficulty to a
	new level by targeting 208 pedestrian-level location classes within a compact urban district
 of only $\sim$70,800\,m\textsuperscript{2},
	with sub-20\,m class separation and severe inter-class visual ambiguity.
	MMS-VPR uniquely integrates images, videos, semantic text, spatial graph structure,
	and a 7-year temporal span---addressing all four critical gaps in existing VPR datasets.
	Our modality ablation experiment confirms systematic gains from multimodal fusion (up to +19.2\% improvement over image-only), and graph-structured modeling provides complementary signal in fine-grained structural subsets. MMS-VPRlib provides a reproducible, standardized evaluation platform supporting 22 baselines
	across 6 VPR datasets~\cite{torii201524, smith2009newcollege, torii2013visual, sunderhauf2013we, kendall2015posenet}.
	Our low-cost, smartphone-based data collection methodology (detailed in Appendix~\ref{app:collection_protocols})can be replicated in other cities,
	supporting future multi-site extensions.
	Promising future directions include space-syntax-aware retrieval (detailed in Appendix~\ref{app:syntax}), GNN-based methods, flow-aware hierarchical
	place search, and cross-city transfer.

	\clearpage

	\bibliographystyle{plainnat}
	\bibliography{mms_vpr_refs}

	\appendix

\clearpage
\appendix

\section{Supplementary Materials for MMS-VPRlib}
\label{app:benchmark}

This section provides additional setup details of hyperparameter configurations for 22 baseline models (RQ1), extended benchamrk result visualizations (RQ1), ablation details (RQ2) hyperparameter configurations, sensitivity analysis results (RQ4), and computational resources for the MMS-VPRlib benchmark platform (Section~\ref{sec:lib}).

\subsection{Computational Resources}
\label{app:compute}

All experiments were conducted on a single consumer laptop (no cloud or
multi-node cluster).

\subsubsection{Hardware Specifications}
\begin{itemize}[itemsep=1pt, topsep=2pt]
    \item \textbf{Laptop}: Lenovo Legion Y7000P IRX9
    \item \textbf{CPU}: Intel Core i7-14700HX (20 cores / 28 threads, 2.1\,GHz base, 33\,MB L3)
    \item \textbf{GPU}: NVIDIA GeForce RTX\,4060 Laptop, 8\,GB GDDR6
    \item \textbf{RAM}: 16\,GB DDR5-5600
    \item \textbf{Storage}: YMTC NVMe SSD, 1\,TB
    \item \textbf{OS}: Windows\,11 Pro 22H2 + WSL2 (Ubuntu 22.04),
          CUDA 12.5, cuDNN\,9
\end{itemize}

\subsubsection{Runtime Analysis}
\label{app:runtime}

Table~\ref{tab:running_time_all_datasets} reports training and testing time
across all five external datasets.
An RTX\,3060/4060 (8\,GB) or equivalent is sufficient to reproduce all results.

\begin{table*}[h]
    \centering
    \caption{Training (Tr) and testing (Te) time across five datasets (minutes).
        Dataset scaling factors relative to New College:
        Tokyo $\times4$, Pittsburgh $\times10.26$,
        Nordland $\times4$, Cambridge $\times1$.}
    \label{tab:running_time_all_datasets}
    \small
    \setlength{\tabcolsep}{5pt}
    \renewcommand{\arraystretch}{1.0}
    \resizebox{\textwidth}{!}{
    \begin{tabular}{l *{10}{c}}
        \toprule
        \multirow{2}{*}{Model}
        & \multicolumn{2}{c}{Tokyo} & \multicolumn{2}{c}{New College}
        & \multicolumn{2}{c}{Pittsburgh} & \multicolumn{2}{c}{Nordland}
        & \multicolumn{2}{c}{Cambridge} \\
        \cmidrule(lr){2-3}\cmidrule(lr){4-5}\cmidrule(lr){6-7}
        \cmidrule(lr){8-9}\cmidrule(lr){10-11}
        & Tr & Te & Tr & Te & Tr & Te & Tr & Te & Tr & Te \\
        \midrule
        CLIP         & 45.9 & 8.1  & 2.1 & 0.4 & 252.8  & 44.6  & 146.2 & 25.8  & 7.7  & 1.3 \\
        BLIP         & 51.0 & 9.0  & 7.2 & 1.3 & 1002.5 & 176.9 & 571.2 & 100.8 & 12.8 & 2.2 \\
        BoQ          & 84.7 & 9.4  & 3.5 & 0.4 & 556.0  & 61.8  & 252.3 & 28.0  & 25.6 & 2.9 \\
        SALAD        & 81.8 & 9.1  & 3.5 & 0.4 & 876.0  & 97.3  & 425.8 & 47.3  & 16.3 & 1.8 \\
        ResNet       & 12.3 & 1.7  & 2.2 & 0.3 & 53.3   & 7.3   & 18.1  & 2.5   & 2.7  & 0.4 \\
        ViT          & 27.1 & 3.7  & 6.6 & 0.9 & 129.2  & 17.6  & 48.6  & 6.6   & 7.6  & 1.0 \\
        CosPlace     & 56.9 & 6.3  & 2.3 & 0.3 & 290.3  & 32.3  & 200.2 & 22.2  & 10.4 & 1.2 \\
        MixVPR       & 54.4 & 6.0  & 2.3 & 0.3 & 290.9  & 32.3  & 159.0 & 17.7  & 10.4 & 1.2 \\
        SFRS         & 55.8 & 6.2  & 2.4 & 0.3 & 295.8  & 32.9  & 165.6 & 18.4  & 10.6 & 1.2 \\
        EigenPlaces  & 54.8 & 6.1  & 2.3 & 0.3 & 288.3  & 32.0  & 159.8 & 17.8  & 10.1 & 1.1 \\
        PatchNetVLAD & 54.6 & 6.1  & 5.0 & 0.6 & 291.7  & 32.4  & 160.1 & 17.8  & 10.4 & 1.1 \\
        R2Former     & 55.2 & 6.1  & 2.3 & 0.3 & 286.2  & 31.8  & 157.0 & 17.4  & 11.7 & 1.3 \\
        \bottomrule
    \end{tabular}}
\end{table*}

\subsection{Hyperparameter Configurations (RQ1)}
\label{app:hyper}

As mentioned in Section~\ref{sec:exp}, we use batch size 32 for most methods (except BLIP: 16, LR: 64). Table~\ref{tab:hyperparams_remaining} details all model-specific hyperparameters for all 22 baselines (excluding common settings).

\FloatBarrier
\begin{table}[h]
    \centering
    \caption{Hyperparameter configurations for all 22 baselines in MMS-VPRlib.}
    \label{tab:hyperparams_remaining}
    \scriptsize
    \setlength{\tabcolsep}{2pt}
    \renewcommand{\arraystretch}{0.95}
    \begin{tabularx}{\columnwidth}{@{}
            >{\raggedright\arraybackslash}p{0.18\columnwidth}
            >{\raggedright\arraybackslash}p{0.20\columnwidth}
            >{\raggedright\arraybackslash}X
            >{\raggedleft\arraybackslash}p{0.14\columnwidth}
            @{}}
        \toprule
        \textbf{Model} & \textbf{Hyperparameter} & \textbf{Description} & \textbf{Value} \\
        \midrule
        ViT~\cite{vit}
        & $d_{\text{embed}}$ & Patch embedding dimensionality & 64 \\
        & $L$ & Number of Transformer layers & 6 \\
        & $H$ & Number of attention heads & 8 \\
        & $d_{\text{mlp}}$ & FFN hidden dimensionality & 128 \\
        \midrule
        SVC~\cite{SVC}
        & $C$ & Regularisation strength & 1.0 \\
        \midrule
        RF~\cite{RF}
        & $N_{\text{trees}}$ & Number of trees & 100 \\
        \midrule
        MLP~\cite{MLP}
        & $d_{\text{hidden}}$ & Hidden layer sizes & \texttt{(256,128)} \\
        & $T_{\text{max}}$ & Maximum iterations & 300 \\
        \midrule
        LR~\cite{LR}
        & $d_{\text{img}}$ & Image branch hidden size & 128 \\
        \midrule
        KNN~\cite{KNN}
        & $K$ & Number of neighbours & 5 \\
        \midrule
        GNB~\cite{GNB}
        & $\epsilon$ & Variance smoothing & $1\times10^{-9}$ \\
        \midrule
        BoQ~\cite{boq}
        & $\mathbb{I}_{\text{pre}}$ & Use pretrained backbone & True \\
        \midrule
        SALAD~\cite{SALAD}
        & $\lambda_{\text{distill}}$ & Distillation loss weight & $5\times10^{-4}$ \\
        \midrule
        BLIP~\cite{li2022blip}
        & $p_{\text{crop}}$ & RandomResizedCrop scale & \texttt{(0.8,1.0)} \\
        \midrule
        CLIP~\cite{radford2021clip}
        & $L_{\text{unfreeze}}$ & Unfrozen Transformer blocks & 1 \\
        & $\phi$ & Feature normalisation (L2) & True \\
        \midrule
        EigenPlaces~\cite{EigenPlaces}
        & $p_{\text{GeM}}$ & GeM pooling exponent & 3.0 \\
        & $d_{\text{proj}}$ & Projection dimension & 256 \\
        & $\lambda_{\text{ortho}}$ & Orthogonality regularisation & $1\times10^{-3}$ \\
        & $s$ & CosFace scale factor & 30.0 \\
        & $m$ & CosFace margin & 0.35 \\
        \midrule
        SFRS~\cite{ge2020self}
        & $r$ & Channel reduction ratio & 16 \\
        & $p_{\text{GeM}}$ & GeM pooling exponent & 3.0 \\
        \midrule
        R2Former~\cite{zhu2023r2former}
        & $d_{\text{embed}}$ & Token embedding dimension & 256 \\
        & $L$ & Transformer encoder layers & 4 \\
        & $H$ & Attention heads per layer & 8 \\
        & $\rho$ & MLP expansion ratio & 4.0 \\
        & $\alpha_{0}$ & Initial R2 fusion weight & 0.5 \\
        \midrule
        PatchNetVLAD~\cite{patchnetvlad}
        & $d_{\text{proj}}$ & Projection dimension & 128 \\
        & $K$ & NetVLAD clusters & 16 \\
        & $\mathcal{G}$ & Multi-scale grid sizes & \texttt{[1,2,3]} \\
        & $R$ & Total regions ($\sum g^{2}$) & 14 \\
        \midrule
        MixVPR~\cite{mixvpr}
        & $d_{\text{proj}}$ & Projection dimension & 256 \\
        & $L_{\text{mix}}$ & Number of Mix blocks & 3 \\
        & $k_{\text{dw}}$ & Depthwise conv kernel size & 7 \\
        & $\rho$ & Channel expansion ratio & 2.0 \\
        & $p_{\text{GeM}}$ & GeM pooling exponent & 3.0 \\
        \midrule
        CosPlace~\cite{CosPlace}
        & $p_{\text{GeM}}$ & GeM pooling exponent & 3.0 \\
        & $d_{\text{feat}}$ & Feature dimension & 512 \\
        & $s$ & CosFace scale factor & 30.0 \\
        & $m$ & CosFace margin & 0.35 \\
        \midrule
        ResNet~\cite{resnet}
        & Depth & Network depth (layers) & 50 \\
        & $d_{\text{feat}}$ & Output feature dimension & 2048 \\
        & $\mathbb{I}_{\text{pre}}$ & ImageNet pretrained weights & True \\
        & $p_{\text{pool}}$ & Global average pooling & True \\

        \midrule
        GCN~\cite{GCN}
        & $d_{\text{hidden}}$ & Hidden dimension & 64 \\
        & $L$ & Number of GCNConv layers & 2 \\
        \midrule
        GAT~\cite{GAT}
        & $d_{\text{hidden}}$ & Hidden dimension & 64 \\
        & $H$ & Attention heads (first layer) & 8 \\
        & $p_{\text{drop}}$ & Dropout rate & 0.6 \\
        & $L$ & Number of GATConv layers & 2 \\
        \midrule
        HGNN~\cite{hgnn}
        & $d_{\text{hidden}}$ & Hidden feature size & 64 \\
        & $L$ & Number of HGTConv layers & 2 \\
        & $H$ & Attention heads per HGTConv & 2 \\
        \midrule
        HGNN+ResNet
        & $S_{\text{img}}$ & Input image size (H,W) & \texttt{(32,32)} \\
        & $B$ & Batch size & 32 \\
        & $L_{\text{block}}$ & BasicBlock layers/stage & \texttt{[2,2,2,2]} \\
        & $C_{\text{in}}$ & Input channels & 1 \\
        \bottomrule
    \end{tabularx}
\end{table}

\subsection{Additional Baseline Benchmark Visualizations (RQ1)}
\label{app:vis}

\begin{figure*}[h]
    \centering
    \includegraphics[width=\textwidth]{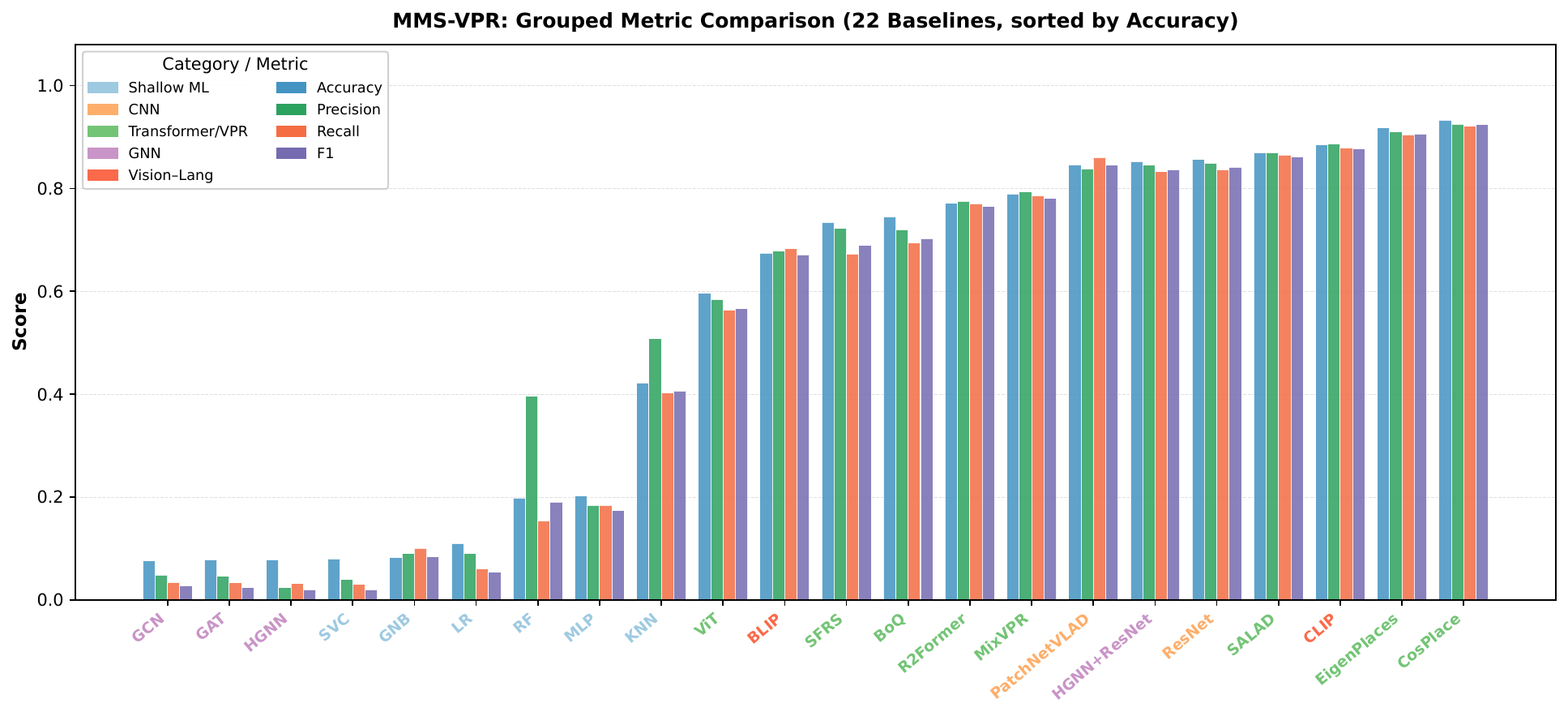}
    \caption{\textbf{Grouped metric comparison across all 22 baselines on MMS-VPR}
        (sorted by Accuracy, low to high).
        Each cluster shows Accuracy, Precision, Recall, and F1 for one model.}
    \label{fig:app_grouped_bar}
\end{figure*}

Figure~\ref{fig:app_grouped_bar} shows a grouped bar chart of the same
results sorted by Accuracy, providing an alternative view of performance
distribution across model families.

\clearpage

\subsection{Modality Ablation: Extended Visualizations (RQ2)}
\label{app:ablation_vis}

\begin{figure*}[h]
    \centering
    \includegraphics[width=\textwidth]{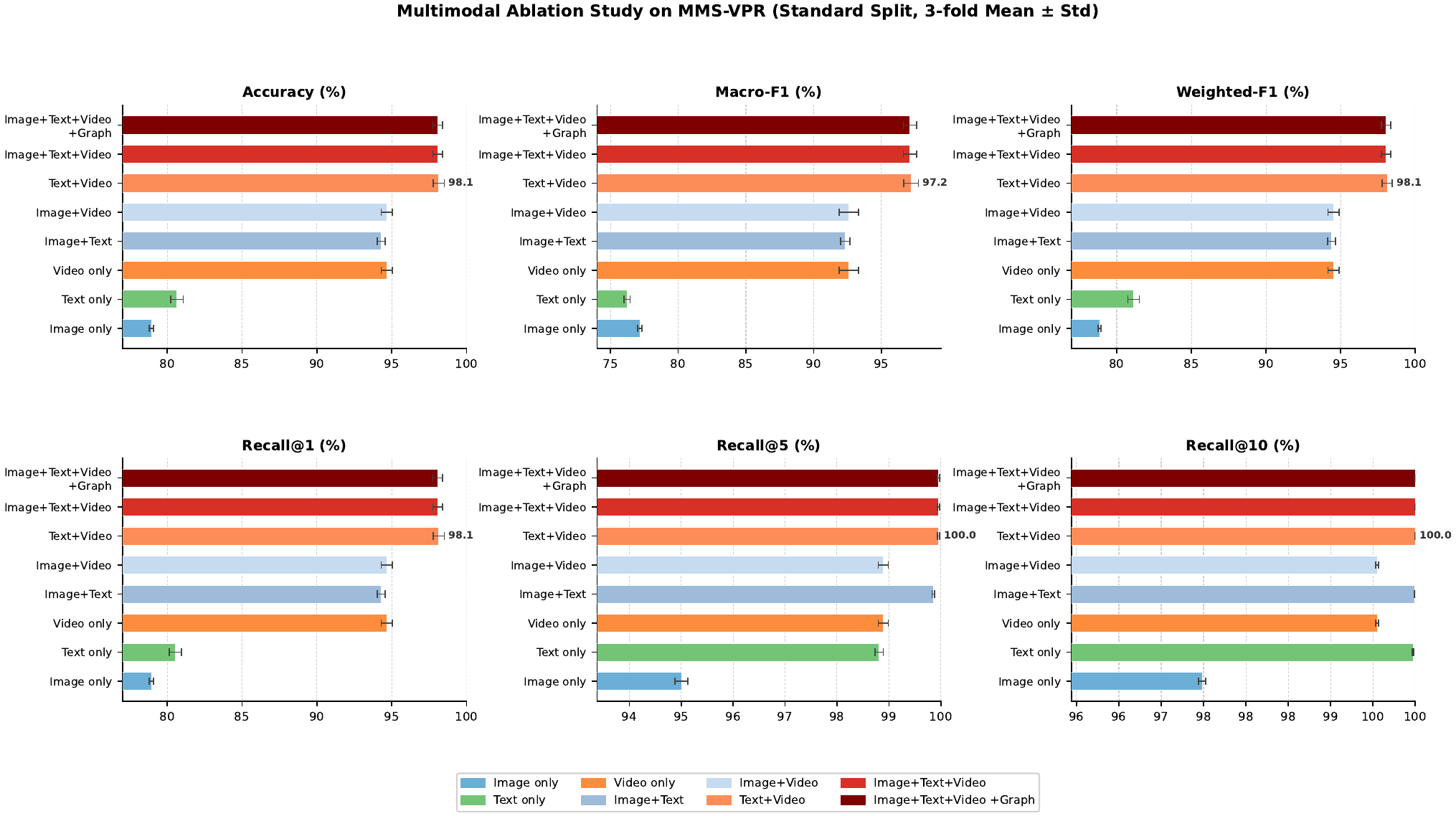}
    \caption{\textbf{Full modality ablation results across six metrics}
        (Accuracy, Macro-F1, Weighted-F1, Recall@1, Recall@5, Recall@10),
        all 8 modality combinations, mean\,$\pm$\,std over 3 splits.}
    \label{fig:app_ablation_main}
\end{figure*}

\begin{figure*}[t]
    \centering
    \includegraphics[width=0.85\textwidth]{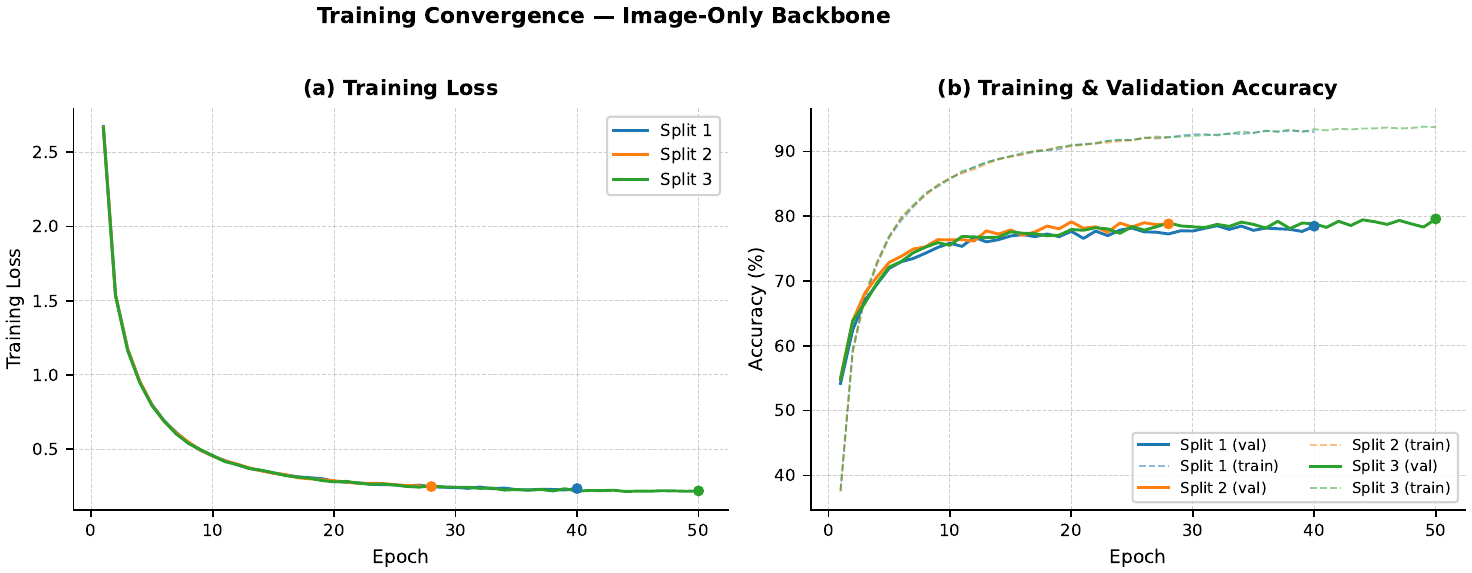}
    \caption{\textbf{Training convergence } of the image-only  backbone
        across 3 random splits.
        (a) Training loss converges smoothly by epoch~25 for all splits.
        (b) Validation accuracy (solid lines) plateaus around 79\%, while
        training accuracy (dashed) continues to rise, indicating the
        performance ceiling discussed in Section~\ref{sec:rq2}.}
    \label{fig:app_convergence}
\end{figure*}

\begin{figure}[t]
    \centering
    \includegraphics[width=0.72\linewidth]{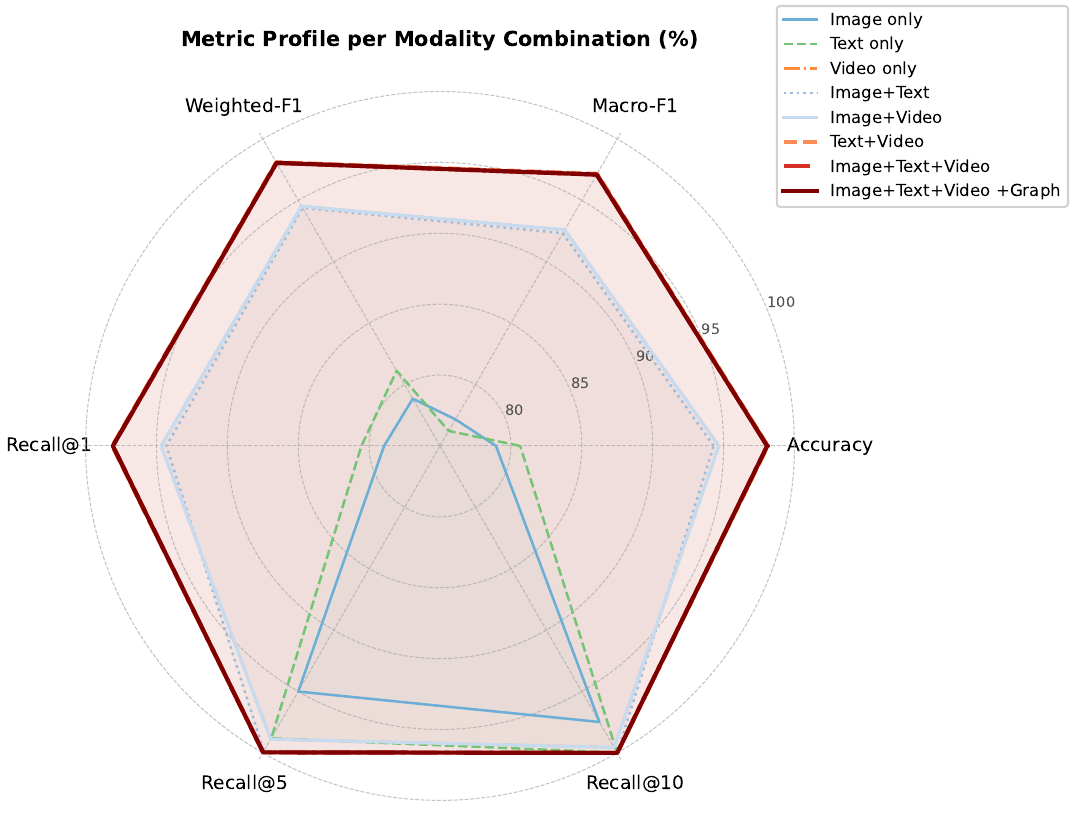}
    \caption{\textbf{Radar profile across six metrics for all modality combinations.}
        Image+Text+Video (and +Graph) achieve the largest enclosed area across
        all metrics. Video-containing combinations dominate on Accuracy and
        Recall@1; Text-containing combinations excel at Recall@5 and Recall@10.}
    \label{fig:app_radar_profile}
\end{figure}

Figures~\ref{fig:app_ablation_main}--\ref{fig:app_radar_profile} provide
extended visualizations for the modality ablation experiment (Section~\ref{sec:rq2}).

\subsection{Ablation Study Hyperparameters (RQ2)}
\label{app:hyper_ablation}

Table~\ref{tab:hyper_ablation} lists the hyperparameters for the modality
ablation study (Section~\ref{sec:rq2}).

\begin{table}[t]
    \centering
    \caption{Hyperparameters for the modality ablation
        (MobileNetV3-Small frozen backbone).}
    \label{tab:hyper_ablation}
    \small
    \setlength{\tabcolsep}{5pt}
    \renewcommand{\arraystretch}{1.05}
    \begin{tabular}{@{} l l l l @{}}
        \toprule
        Component & Hyperparameter & Description & Value \\
        \midrule
        \multirow{2}{*}{Visual}
        & Backbone & CNN architecture & MobileNetV3-Small \\
        & Backbone weights & Pretrained (frozen) & ImageNet \\
        Text   & Encoder & Language model (frozen) & BERT-base \\
        \midrule
        \multirow{2}{*}{Split}
        & Train/Val/Test & Per-class stratified & 80\%/10\%/10\% \\
        & Num splits & Random seeds & 3 \\
        \midrule
        \multirow{5}{*}{Training}
        & Optimizer & --- & Adam \\
        & Learning rate & --- & $1\times10^{-3}$ \\
        & Batch size & --- & 64 \\
        & Max epochs & --- & 50 \\
        & Early stopping & Val accuracy patience & 8 epochs \\
        \midrule
        \multirow{2}{*}{Fusion}
        & Weight init & Per-modality scalar & Uniform \\
        & Weight constraint & Simplex projection & $\sum w=1$ \\
        \bottomrule
    \end{tabular}
\end{table}

\subsection{Code Repository}
\label{app:code_repository}

Complete implementation is available at \url{https://github.com/yiasun/MMS-VPRlib}.
The repository includes:
\begin{itemize}[itemsep=1pt, topsep=2pt]
    \item Jupyter notebooks for all 22 baseline models
    \item Conda environment files for full reproducibility
    \item Training and evaluation scripts for all supported datasets
    \item Data loading utilities and preprocessing pipelines
\end{itemize}
Each notebook is self-contained and covers data loading, preprocessing,
training, and evaluation.
Exact commands and environment specifications are provided in the repository
README.

\section{Further Details of MMS-VPR Dataset}
\label{app:dataset}

\subsection{Related Work in Visual Place Recognition}
\label{app:related_work}

Visual Place Recognition (VPR) in urban street scenes has emerged as a
critical task in autonomous navigation and geo-localisation.
This subsection reviews representative datasets in terms of collection
modalities, environmental scope, and sensor setups.

\noindent\textbf{Multisensor Robotic Platforms.}
The University of Michigan North Campus dataset~\cite{University_of_Michigan}
integrates omnidirectional and planar LiDAR, IMU, gyroscope, and RTK GPS
for long-term SLAM and appearance-based place recognition over 15 months.

\noindent\textbf{Mobile Vehicle and Camera-Rig Systems.}
Chen et al.~\cite{City_Scale} constructed a 1.7M-image dataset using
panoramic cameras and GPS.
NYU-VPR~\cite{NYU_VPR} employs smartphone cameras on fleet cars capturing
front and side views across day/night cycles.

\noindent\textbf{Crowdsourced and Time-Machine Imagery.}
Mapillary Vistas~\cite{Mapillary_Vistas} and Street-Level
Sequences~\cite{Mapillary_Street_Level} utilise user-contributed imagery
from smartphones and dashcams.
GSV-Cities~\cite{GSV_Cities} offers 560K images spanning 14 years across
40 cities, suitable for VPR under appearance change and domain shift.

\noindent\textbf{Synthetic and Archival Imagery.}
AmsterTime~\cite{yildiz2022amstertime} integrates archival scans with modern street views
for cross-domain VPR evaluation.
HPointLoc~\cite{HPointLoc} simulates indoor scenes via RGB-D rendering from
Matterport3D.

\noindent\textbf{Cross-View and Multi-Modal Datasets.}
CV-Cities~\cite{CV_Cities} and SpaGBOL~\cite{SpaGBOL} bridge
ground-to-satellite geo-localisation using structured road-junction graph
annotations.

\noindent\textbf{Event-Based and Neuromorphic Imaging.}
NYC-Event-VPR~\cite{NYC_Event_VPR} uses a Prophesee HD event camera coupled
with RGB and GPS, covering 260\,km of New York City under diverse conditions.

\noindent\textbf{Urban Street-Scene VPR Methods.}
MS-MixVPR~\cite{quach2024ms} presents multi-scale feature fusion combining
ResNet50 with a FeatureReuse module.
MeshVPR~\cite{berton2024meshvpr} leverages synthetic city mesh reconstructions.
GICNet~\cite{wu2024gicnet} uses the Shuffle Channel Attention module and
Global Information Aggregator for day-night robustness.

\subsection{Site Selection Rationale and Characteristics}
\label{app:site_details}

\subsection{Site Location and Spatial Layout}
\label{app:site location}
\begin{figure}[h]
	\centering
	\includegraphics[width=0.8\linewidth]{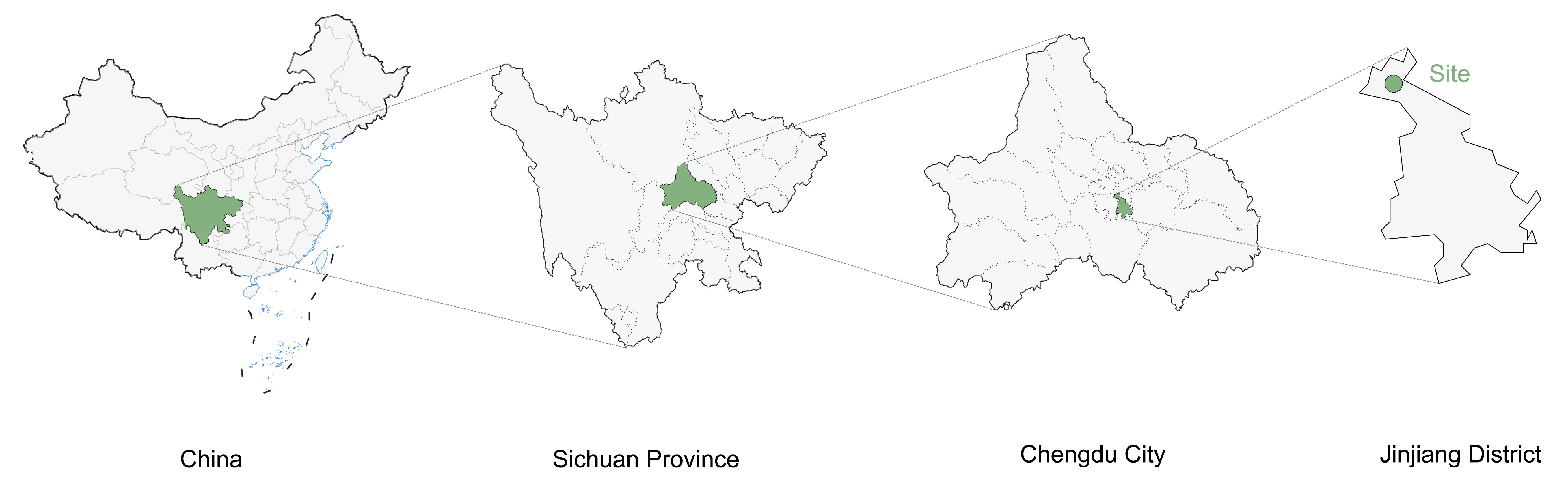}
	\caption{Geo-location of the proposed dataset site (Chengdu Taikoo Li).}
	\label{Site}
\end{figure}

\begin{figure}[h]
	\centering
	\includegraphics[width=0.75\linewidth]{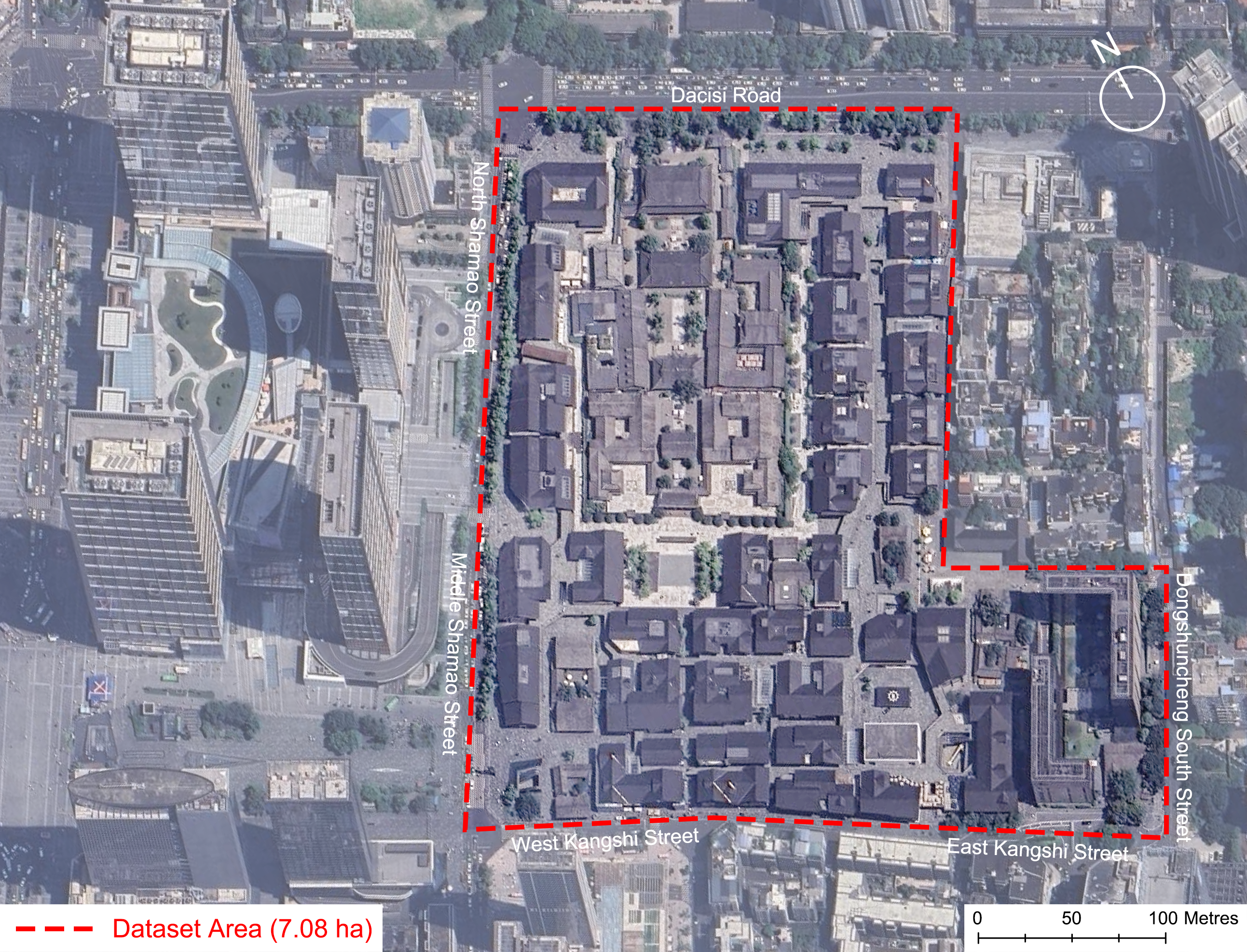}
	\caption{Boundary and spatial layout of the pedestrian streets within the site.}
	\label{Satellite}
\end{figure}

As shown in Figure~\ref{Site}, Chengdu Taikoo Li is located at the central business district (CBD), the Jinjiang district of Chengdu City, the capital of Sichuan Province, China. The development covers approximately 70,800\,m\textsuperscript{2} and was jointly developed by Swire Properties and Sino-Ocean Group. As illustrated in Figure~\ref{Satellite}, the site comprises more than 40 two-to-three storey individual buildings interconnected by a network of open-air pedestrian pathways that integrate seamlessly with surrounding urban streets including Dacisi Road, Kangshi Street, and Shamao Street.

\subsubsection{Advantages Over Indoor Mall Datasets}

Previous urban commercial-environment VPR datasets primarily focus on interior spaces of enclosed indoor malls. For instance, \citet{lee2021large} proposed a localization dataset of indoor shopping mall spaces in Seoul, South Korea. However, these VPR datasets present several limitations:

\begin{itemize}
	\item \textbf{Low visual diversity:} Interior mall spaces typically feature controlled, artificial lighting conditions with minimal temporal variation, limiting the model's ability to recognize identical places under different lighting environments.

	\item \textbf{Insufficient spatial uniqueness:} Interior commercial spaces, particularly brand stores (e.g., Louis Vuitton, Nike, Gucci), often employ standardized design languages across different locations worldwide. This lack of uniqueness impedes the development of place recognition algorithms with global applicability.

	\item \textbf{Limited environmental complexity:} Indoor environments lack natural variations in lighting, weather, and occlusion patterns that characterize real-world urban settings.

	\item \textbf{Restricted generalization:} Models trained on interior mall datasets may struggle to generalize to the diverse conditions found in outdoor urban environments.
\end{itemize}

MMS-VPR addresses these limitations by focusing on outdoor pedestrian streets within a precinct mall setting. These spaces feature:

\begin{itemize}
	\item \textbf{Environmental variability:} Comprehensive surrounding urban context features, skylines, natural lighting conditions, and varying pedestrian flows that ensure absolute and global uniqueness of place recognition results.

	\item \textbf{Semantic richness:} Varied functions including shopping, dining, leisure, art, and cultural spaces that provide diverse contextual cues for recognition.

	\item \textbf{Structural openness:} Long sightlines, rich occlusion patterns, and varied viewpoints that challenge and improve model generalization.

	\item \textbf{Temporal robustness:} Environmental variation across day/night cycles and seven years of social media imagery, providing a more robust training foundation for real-world place recognition algorithms.
\end{itemize}

These features make it particularly suitable for examining the performance of place recognition algorithms in real-world urban scenarios with complex pedestrian dynamics, addressing a significant gap in existing place recognition datasets.

\subsection{Data Collection Protocols}
\label{app:collection_protocols}

\subsubsection{Device Specifications}
\label{app:devices}

Table~\ref{tab:device-data-format} summarises the consumer smartphones used
for all data collection.

\begin{table}[h]
	\caption{The main device and deliverable raw data format of the proposed dataset}
	\label{tab:device-data-format}
	\centering
	\begin{tabular}{llll}
		\toprule
		Device & Model & Output Image Format & Output Video Format \\
		\midrule
		Smartphone & iPhone XS Max & 4032×3024 & 1920×1080, 30fps \\
		Smartphone & iPhone 11 Pro Max & 4032×3024 & 1920×1080, 30fps \\
		\bottomrule
	\end{tabular}
\end{table}

A handheld gimbal was used during video recording to improve stability.

\subsubsection{Four-Direction Coverage}
\label{app:four_directions}

\begin{figure}[t]
    \centering
    \includegraphics[width=0.8\linewidth]{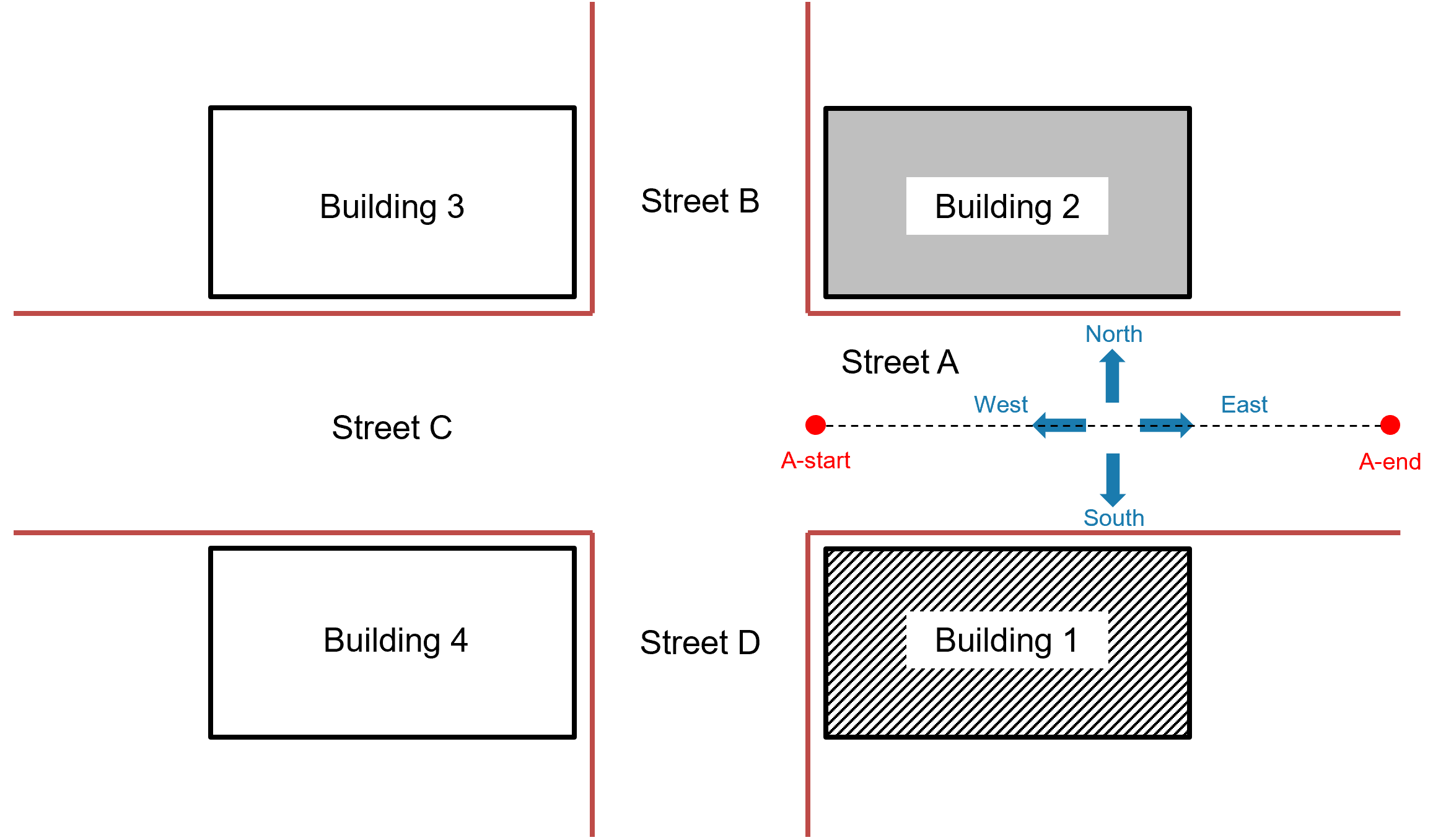}
    \caption{Collection guidelines for four-direction visual coverage on a
        representative street segment.}
    \label{Fig4}
\end{figure}

Figure~\ref{Fig4} illustrates the four-direction collection framework.
For each street segment, the collector walks the full length while facing
each of the four cardinal directions in sequence (E, N, W, S), capturing
both pedestrian-level architectural features and lateral building facades.

\subsubsection{Dual-Perspective Capture}
\label{app:two_perspectives}

\begin{figure}[h]
	\centering
	\includegraphics[width=0.9\linewidth]{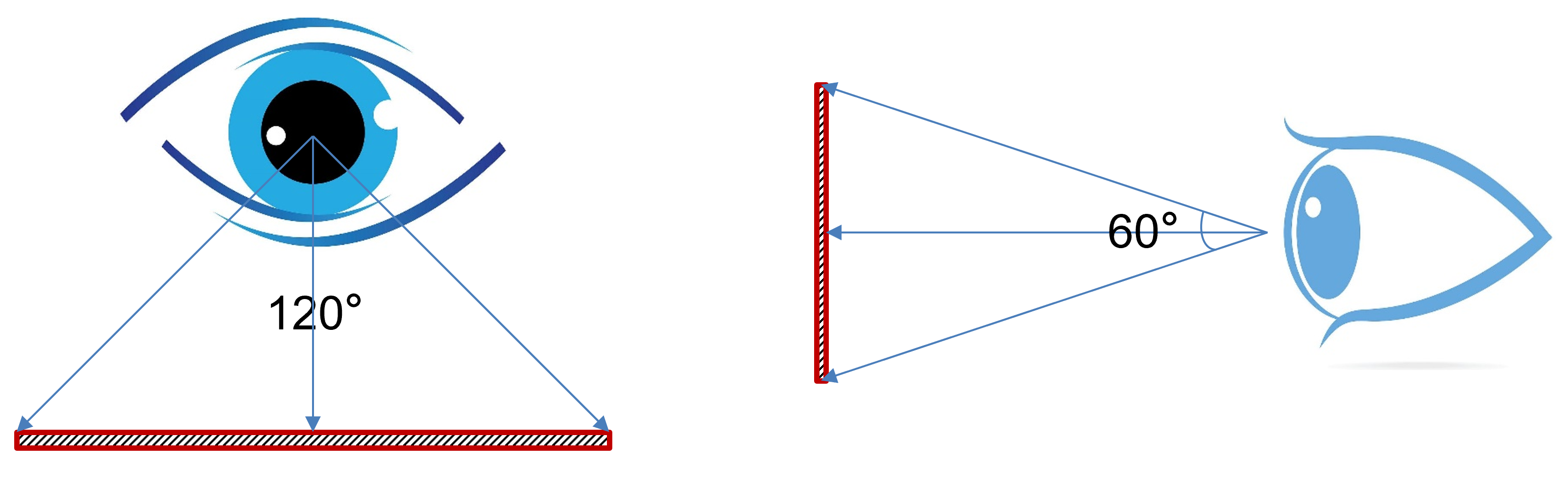}
	\caption{Illustration of the effective field of view (FOV) of human eyes.}
	\label{Fig6}
\end{figure}

Research by \cite{wang2013visual, sanders1993human} indicates that the effective field of human vision is approximately 120° horizontally and 60° vertically \cite{atchison2023optics}, as illustrated in Figure~\ref{Fig6}. Common smartphone cameras offer similar fields of view. For example, the iPhone 11 Pro Max provides an ultra-wide (0.5x) camera with approximately 120° horizontal and 90° vertical FOV, while its wide (1x) camera delivers around 65° horizontal and 46° vertical FOV \cite{apple2019iphone11promax, dxomark2019}. Table~\ref{tab:fov-comparison} summarizes these comparative specifications.

For urban environments with tall buildings, capturing visual features from different perspectives is essential. While a single horizontal perspective may suffice for low-density, low-rise urban areas, high-rise environments require additional angles to comprehensively capture spatial characteristics.

We employ two shooting angles to simulate human viewing behaviors in high-rise urban settings:
\begin{enumerate}[leftmargin=1.5em, itemsep=2pt, parsep=0pt, topsep=2pt]
	\item 0° horizontal angle: Representing the standard forward-looking perspective when walking — to capture typical eye-level architectural features.
	\item 45° upward angle: Simulating the natural head tilt when observing tall buildings — to capture upper-level urban context and building skylines.
\end{enumerate}

These two perspectives, combined with the four directional views, ensure comprehensive capture of panoramic visual features. Additionally, we use both 0.5x (ultra-wide) and 1x (standard) camera modes to account for the variety of device settings used in social media content creation.

\begin{table}[h]
	\caption{Comparison between the Smartphone Cameras (Model: iPhone 11 Pro Max) and the human eyes in terms of the field of views (FOV).}
	\label{tab:fov-comparison}
	\centering
	\begin{tabular}{lllll}
		\toprule
		Apparatus & Mode & Lens & Horizontal FOV & Vertical FOV \\
		\midrule
		\multirow{2}{*}{Smartphone Cameras} & 0.5x & Ultra-Wide Lens Camera & $\sim$120$^\circ$ & $\sim$90$^\circ$ \\
		& 1x & Wide Lens Camera & $\sim$65$^\circ$ & $\sim$46$^\circ$ \\
		Human Eyes & N/A & N/A & $\sim$120$^\circ$ & $\sim$60$^\circ$ \\
		\bottomrule
	\end{tabular}
\end{table}

\FloatBarrier

\subsubsection{Temporal Coverage Strategy}
\label{app:temporal_diversity}

\begin{figure}[h]
    \centering
    \includegraphics[width=\linewidth]{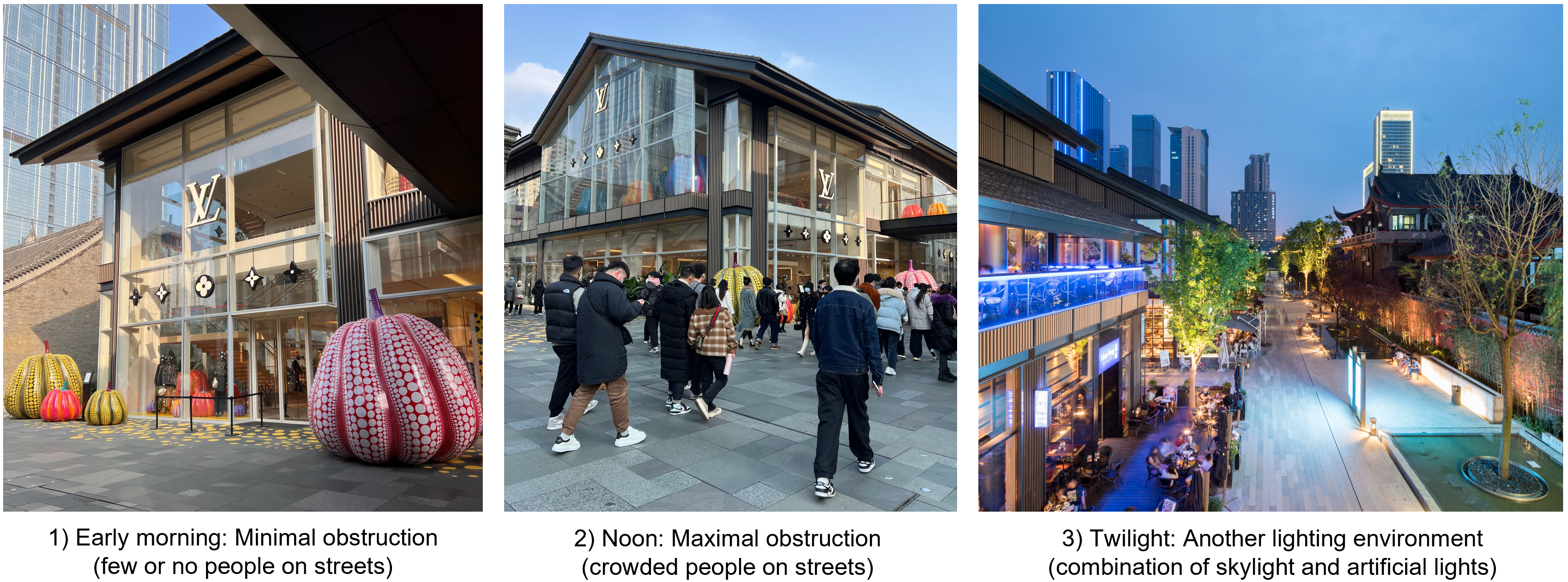}
    \caption{Street scenes during three daytime intervals:
        early morning (minimal occlusion, full daylight),
        noon (heavy pedestrian occlusion), and
        twilight (mixed natural and artificial lighting).}
    \label{Fig9}
\end{figure}

Data collection balanced daytime (07:00--17:00) and nighttime (18:00--22:00)
coverage, with three fine-grained daytime intervals distinguished by
crowd density and lighting conditions.
Figure~\ref{Fig9} illustrates these three intervals;
Table~\ref{tab:visual-features-time} summarises their characteristics.

\begin{table}[h]
	\caption{Visual feature characteristics across temporal periods.}
	\label{tab:visual-features-time}
	\centering
	\begin{tabular}{lllll}
		\toprule
		Period & Sub-period & Time Frame & \makecell[l]{Pedestrian Occlusion on \\Architectural Features} & \makecell[l]{Lighting\\ Environment} \\
		\midrule
		\multirow{3}{*}{Daytime}
		& Early morning & 7am--9am & \makecell[l]{Minimal obstruction\\ (few or no people on streets)} & Daylight \\
		& Noon & 12pm--2pm & \makecell[l]{Moderate - maximal obstruction\\ (crowded people on streets)} & Daylight \\
		& Twilight & 5pm--6pm & \makecell[l]{Moderate - maximal obstruction\\ (crowded people on streets)} & \makecell[l]{Daylight + \\Artificial lights} \\
		Nighttime & N/A & 6pm--10pm & \makecell[l]{Moderate - maximal obstruction\\ (crowded people on streets)} & Artificial lights \\
		\bottomrule
	\end{tabular}
\end{table}

\subsection{Dataset Structure and Encoding}
\label{app:dataset_structure}

\subsubsection{Geographic Distribution}

\begin{figure}[t]
    \centering
    \includegraphics[width=0.9\linewidth]{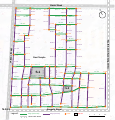}
    \caption{Geographic distribution of all 208 locations in MMS-VPR.}
    \label{fig:geo_distribution_all_places}
\end{figure}

Figure~\ref{fig:geo_distribution_all_places} visualises the geographic
distribution of all 208 locations across the precinct.

\subsubsection{Spatial Encoding Methodology}
\label{app:encoding_methodology}

\begin{figure}[h]
    \centering
    \includegraphics[width=0.6\linewidth]{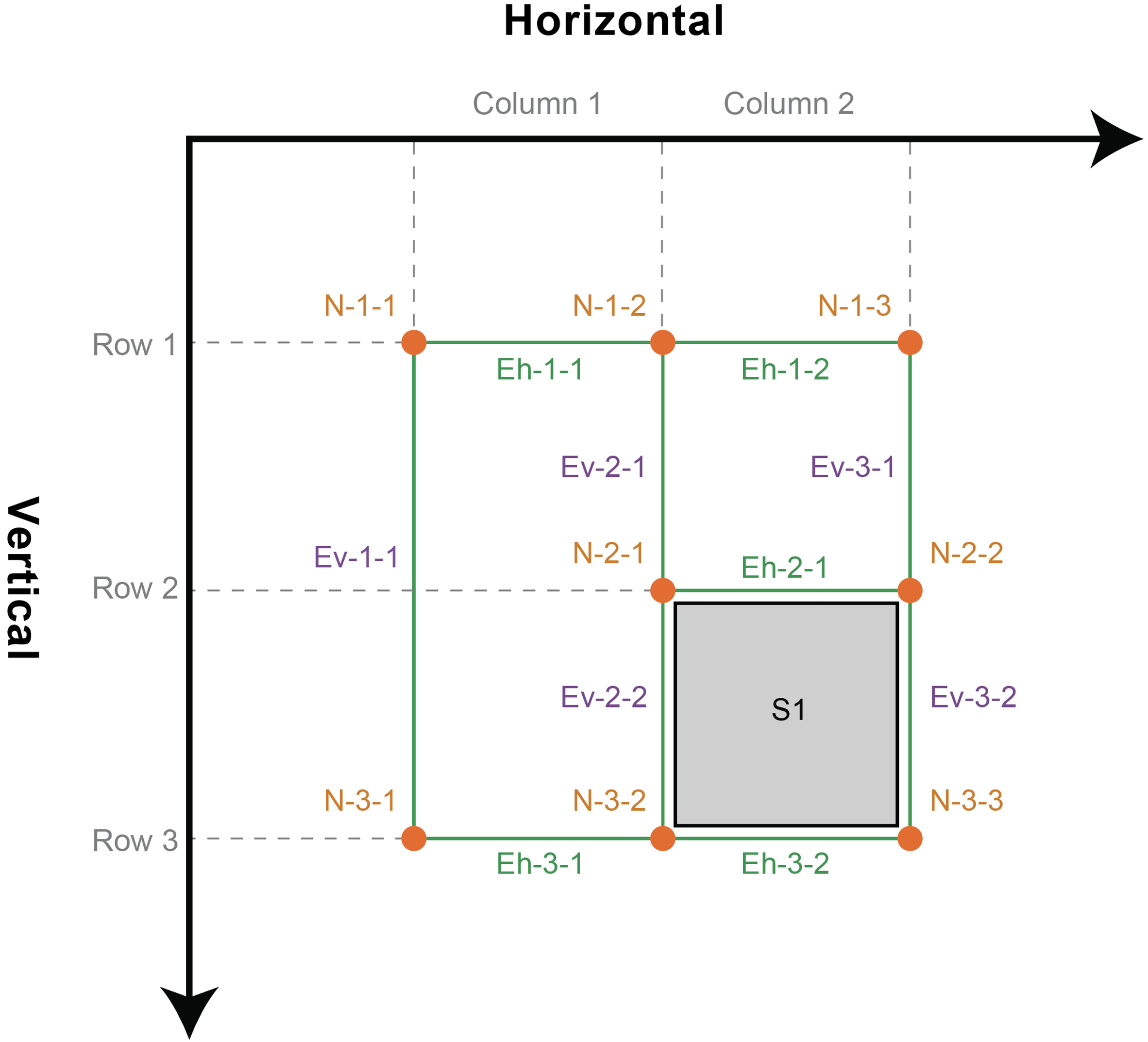}
    \caption{Encoding examples for different location types in MMS-VPR.}
    \label{Fig13}
\end{figure}

Each of the 208 locations receives a unique identifier preserving geographic
position and topological relationships (Figure~\ref{Fig13}):
\begin{itemize}[itemsep=1pt, topsep=2pt]
    \item \textbf{Horizontal streets}: ``Eh-$i$-$j$'' (row $i$, column $j$)
    \item \textbf{Vertical streets}: ``Ev-$n$-$m$'' (column $n$, row $m$)
    \item \textbf{Intersections}: ``N-$i$-$j$''  (row $i$, column $j$)
    \item \textbf{Open squares}: ``S-$r$'' (ranked by area)
\end{itemize}

\subsection{Space Syntax Annotations}
\label{app:syntax}

Traditional VPR datasets provide only visual appearance and coordinates, overlooking spatial
	configuration context that shapes human navigation~\cite{hillier1993natural,peponis1997description}.
	We integrate \emph{space syntax} theory~\cite{hillier1984social,hillier2007space} to enrich
	annotations with spatial configuration metrics, enabling future research in:
	(1) topological context modeling;
	(2) flow-aware retrieval;
	(3) hierarchical place search;
	(4) cross-city transfer.

\subsubsection{Metric Computation}

	Space syntax comprises two main metrics.
	\emph{Integration} (global accessibility; higher = centrally located, easily accessible streets):

	\begin{equation}
		\text{Integration}_i = \frac{k-2}{2\!\left(\frac{1}{k-1}\sum_{j\neq i}d(i,j)-1\right)}
	\end{equation}

	where $d(i,j)$ is shortest-path distance between streets $i$ and $j$.
	\emph{Betweenness} (through-movement potential; higher = primary route, heavier pedestrian flow):

	\begin{equation}
		\text{Betweenness}_i = \sum_{\substack{j\neq i \\ k\neq i,\,k\neq j}}
		\frac{\sigma_{jk}(i)}{\sigma_{jk}}
	\end{equation}

	where $\sigma_{jk}(i)$ counts shortest paths passing through street $i$.
	We employ both \emph{angular distance} $d_a = \sum_{\text{turns}}\theta/90^{\circ}$
	and \emph{weighted distance} $d_w = 0.5\cdot\text{norm}(d_e)+0.5\cdot\text{norm}(d_a)$,
	yielding four properties per street (angular integration, weighted integration, angular
	betweenness, weighted betweenness).

	\subsubsection{Visualizations and Interpretation}

\begin{figure}[h]
    \centering
    \begin{subfigure}[t]{0.48\linewidth}
        \centering
        \includegraphics[width=\linewidth]{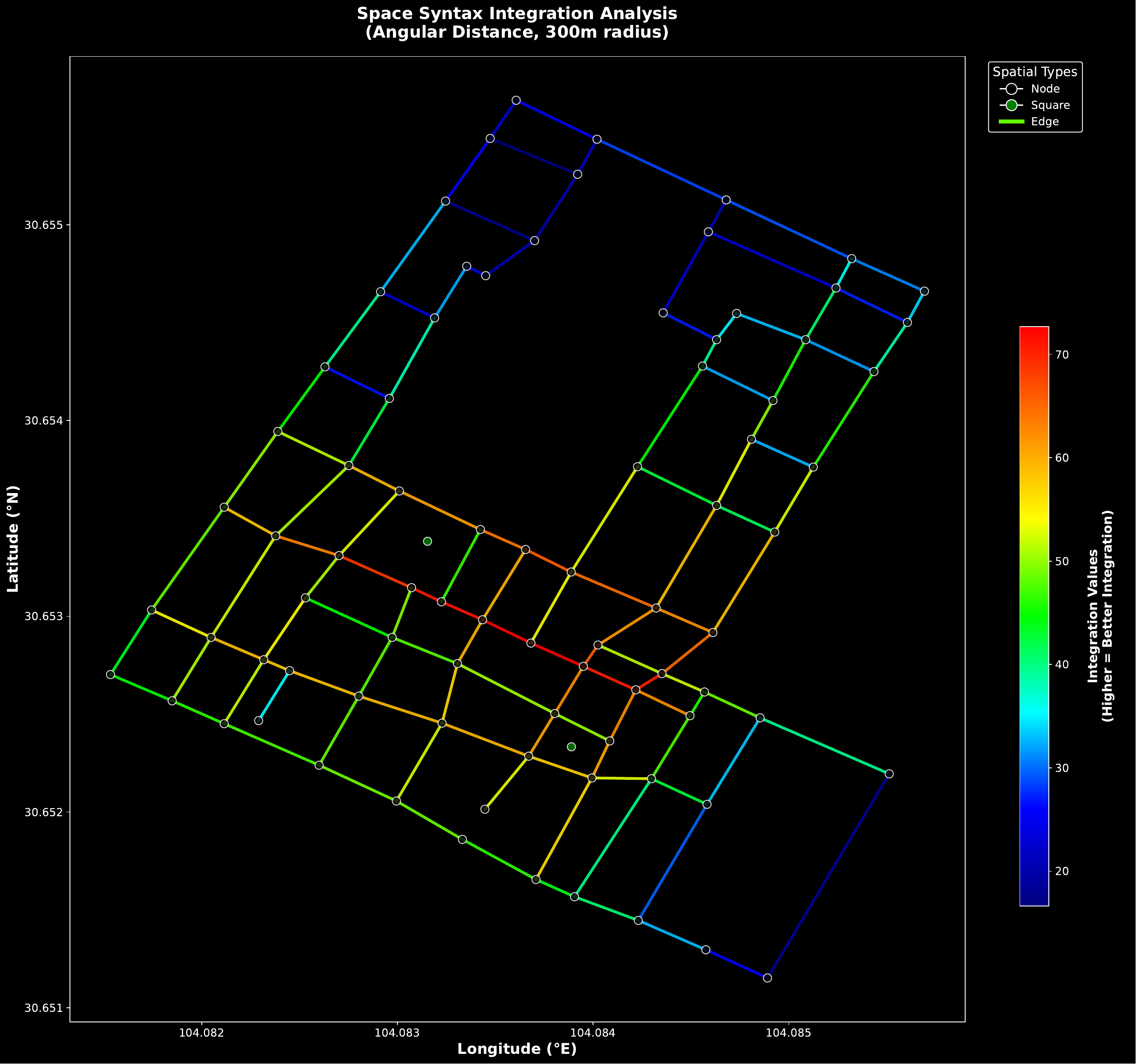}
        \caption{Integration metric. Warmer = higher accessibility.}
        \label{fig:integration_app}
    \end{subfigure}\hfill
    \begin{subfigure}[t]{0.48\linewidth}
        \centering
        \includegraphics[width=\linewidth]{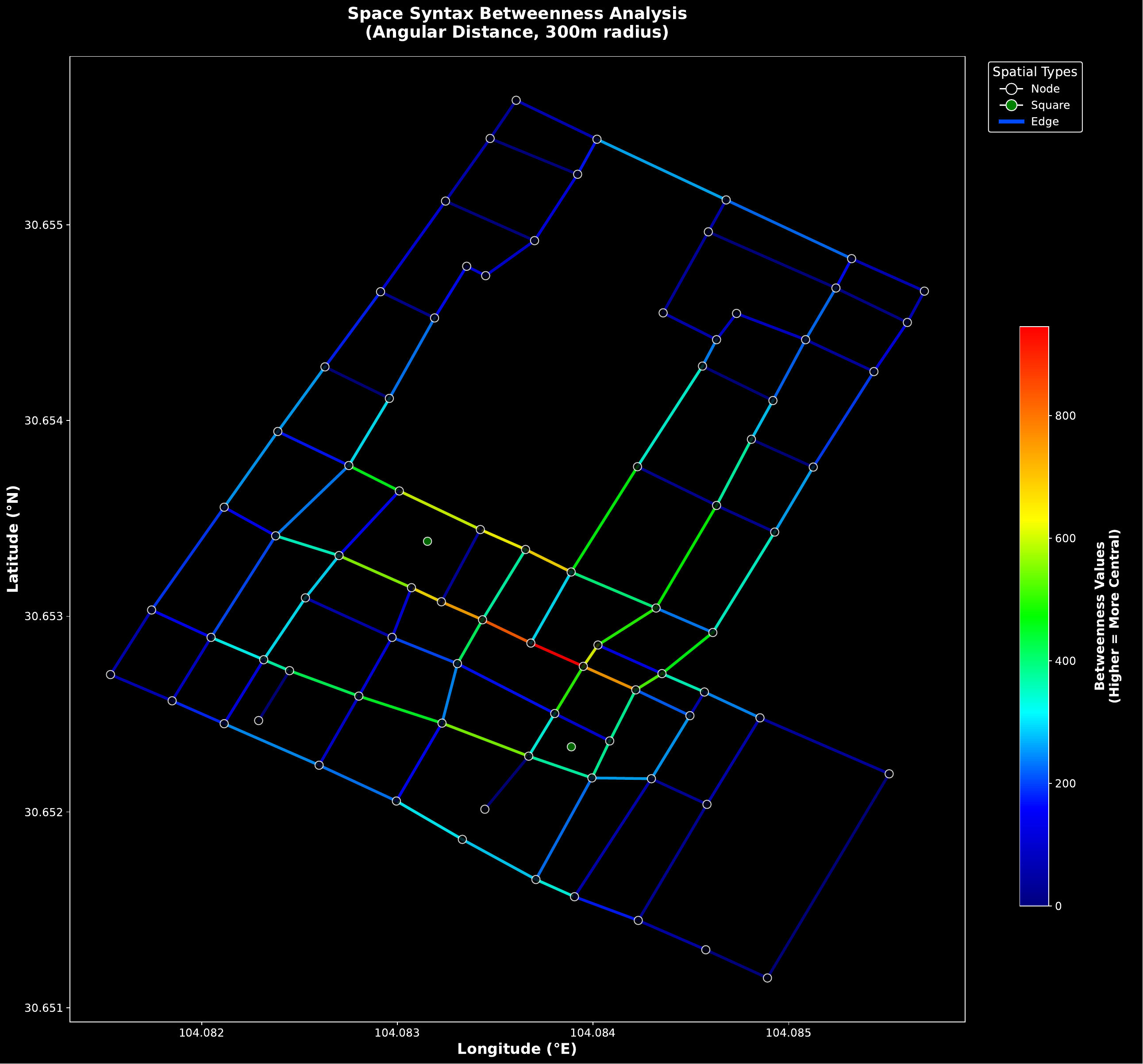}
        \caption{Betweenness metric. Warmer = larger flows.}
        \label{fig:betweenness_app}
    \end{subfigure}
    \caption{Space syntax metrics on the MMS-VPR pedestrian network.}
    \label{fig:syntax_maps}
\end{figure}

	Figures~\ref{fig:integration_app} and~\ref{fig:betweenness_app} visualize
	integration and betweenness across the MMS-VPR network.
	Warm colors (high values) identify central corridors that attract pedestrian movement
	and serve as primary navigation routes; cool colors (low values) indicate peripheral alleys.
	This spatial hierarchy aligns with observed pedestrian flow patterns, validating the
	utility of space syntax for predicting movement potential.

	\subsubsection{Planned Future Experiments}

	Space syntax annotations are dataset-level enrichments, not current benchmark inputs.
	Concrete next steps enabled by these annotations include:
	(1) \emph{Flow-weighted graph retrieval}---using betweenness as edge weights in GNN;
	(2) \emph{Hierarchical search}---using integration values to stratify candidate locations
	before fine-grained matching;
	(3) \emph{Context-aware query routing}---weighting model confidence by spatial
	accessibility scores.
	These experiments require specialized architectures beyond the current benchmark scope
	and are left for future work.

\FloatBarrier

	\subsection{Spatiotemporal Distribution of Social Media Data}
	\label{app:spatiotemporal}

\begin{figure}[h]
		\centering
		\includegraphics[width=0.85\linewidth]{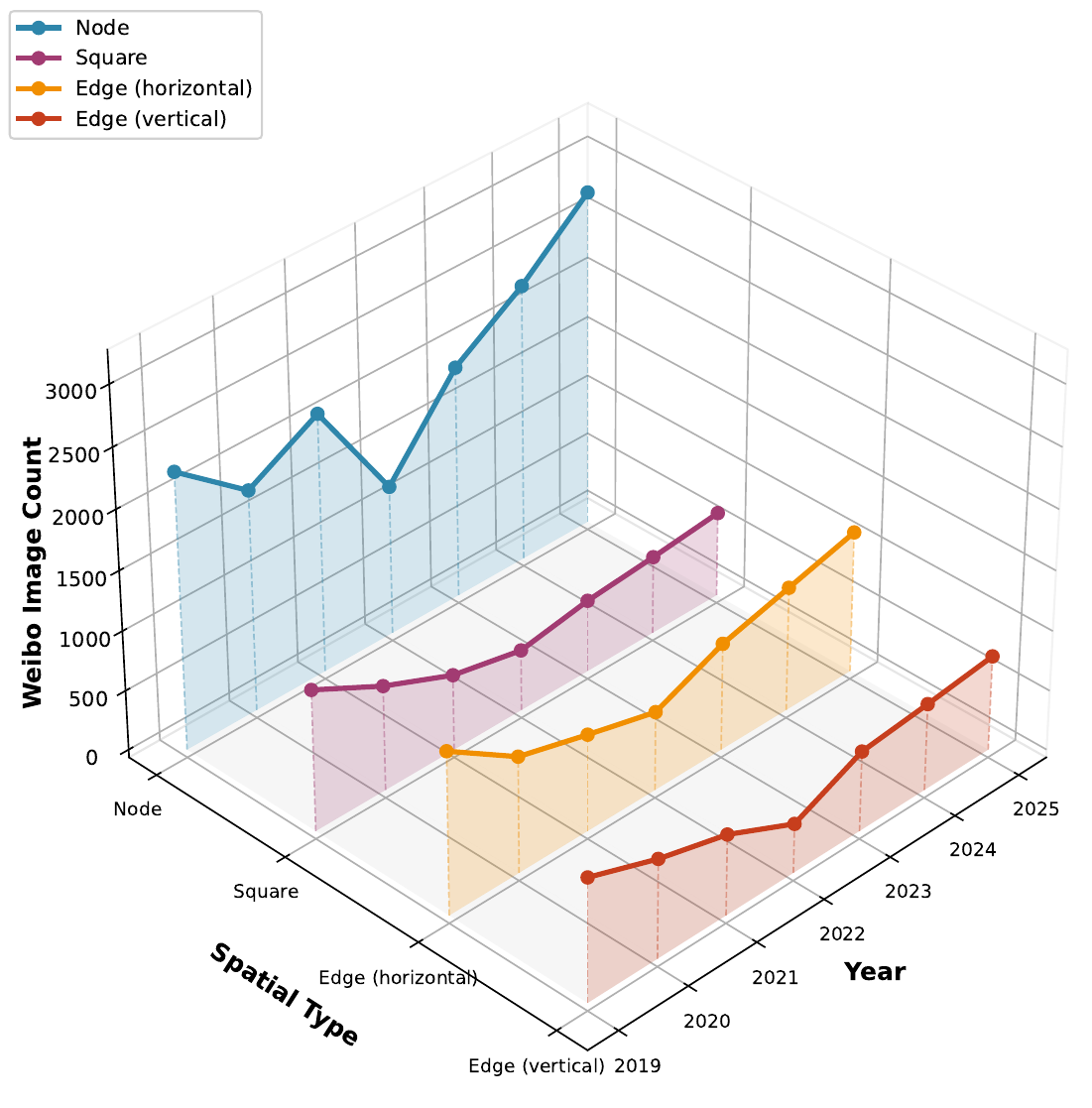}
		\caption{3D spatiotemporal distribution of 31,954 social media images across 4 spatial
			types and 7 years (2019--2025). The 2019-2022 decline reflects COVID-19 influences.}
		\label{fig:spatiotemporal}
	\end{figure}

	Figure~\ref{fig:spatiotemporal} visualizes the spatiotemporal distribution of 31,954
	Weibo images across 7 years (2019--2025) and 4 spatial types.
	Notable observations:
	(1)~Node locations (street intersections) consistently attract the most social media
	activity, reflecting pedestrians' tendency to photograph at decision points;
	(2)~A clear dip in 2022 reflects COVID-19-related restrictions on public gatherings and
	commercial activity in China;
	(3)~Activity recovered to above pre-pandemic levels by 2025 (5,459 images), supporting
	the long-term temporal modeling value of this dataset.

\subsection{Dataset Access and Usage}
\label{app:dataset_usage}

The complete MMS-VPR dataset is publicly available at:\\
\url{https://huggingface.co/datasets/Yiwei-Ou/MMS-VPR}

Data are provided preprocessed at standardised resolution
(images: 256$\times$192; video frames: 256$\times$144), totalling
approximately 5\,GB.

\noindent\textbf{Download:}
\begin{verbatim}
wget https://huggingface.co/datasets/Yiwei-Ou/\
MMS-VPR/resolve/main/MMS-VPR.tar.gz
tar -xzf MMS-VPR.tar.gz
\end{verbatim}

The repository includes annotation schemas, data organisation documentation,
and example usage scripts.

\noindent\textbf{Train/test split.}
Splits are generated per location (per-class stratified) to ensure every
class appears in both sets, eliminating contamination from geographically
adjacent images.
Each class is shuffled and split 80\%/20\%.
The split script (\texttt{split\_generation.py}) is available in the
GitHub repository.

\subsection{Data Governance and Licensing}
\label{app:licensing}

All data in MMS-VPR are released under the
\textbf{Creative Commons Attribution--NonCommercial--ShareAlike 4.0
International (CC\,BY-NC-SA\,4.0)} license.
This applies uniformly to both field-collected and social media imagery,
ensuring non-commercial use, privacy protection, and prevention of dataset
misuse.

\noindent\textbf{Field-collected data (78,575 images, 2,527 videos).}
Collected with written permission from Chengdu Taikoo Li management for
non-commercial academic research.
Data collection was confined to publicly accessible outdoor pedestrian areas.

\noindent\textbf{Weibo-sourced data (31,954 images).}
All Weibo images are:
(1) publicly posted content on the Weibo platform;
(2) anonymised via automated face detection/pixelation and license plate masking;
(3) resolution-limited to a maximum long edge of 256\,pixels, making individual
identification technically infeasible;
(4) restricted to \textbf{non-commercial academic use} under CC\,BY-NC-SA\,4.0.

As a fallback for practitioners requiring unrestricted redistribution,
we provide precomputed CLIP ViT-B/32 embeddings (512-dim) for the Weibo
subset, preserving multimodal utility without redistribution concerns.
The field-collected subset alone constitutes a self-contained, fully
functional multimodal VPR benchmark.

\section{Ethics Statement}
\label{app:ethics}

Our research fully adheres to the NeurIPS Code of Ethics.

\noindent\textbf{Written Institutional Permission.}
All field data were collected following written permission from Chengdu
Taikoo Li management, authorising non-commercial academic data collection
and publication.
Collection was confined to publicly accessible outdoor pedestrian areas.

\noindent\textbf{Privacy Protection.}
All images and videos---field-collected and Weibo-sourced---underwent:
(1) automated face detection and pixelation;
(2) automated license plate detection and masking;
(3) resolution standardisation to a maximum long edge of 256\,pixels
(images: 256$\times$192; video frames: 256$\times$144).
At 256\,px, individual identification is technically infeasible even without
face blurring, consistent with Mapillary Vistas~\cite{Mapillary_Vistas}.
All media were manually reviewed post-processing to verify complete anonymisation.

\noindent\textbf{Public Space Context.}
Data were captured exclusively in publicly accessible outdoor commercial
spaces where pedestrians naturally expect to be observable.
No private, restricted, or indoor spaces were captured.
Collection used handheld consumer devices at natural pedestrian eye level,
avoiding surveillance-style elevated viewpoints.

\noindent\textbf{Dual-Use Risks and Mitigation.}
The dataset's fine-grained spatial structure could theoretically be misused
for location-based profiling.
We mitigate this through: (1) face/plate anonymisation, (2) 256\,px
resolution limit, (3) CC\,BY-NC-SA\,4.0 licensing prohibiting commercial use,
and (4) usage guidelines in the Hugging Face data card explicitly
discouraging downstream surveillance applications.
MMS-VPR is designed to advance research in navigation, AR, and robotics---not
surveillance.

\section{Broader Impacts}
\label{app:broader}

MMS-VPR is designed to advance research in pedestrian-level navigation,
augmented reality, urban analytics, and autonomous systems.
The dataset's fine-grained spatial structure and multimodal richness
enable development of more robust, real-world-deployable place recognition
systems.

Potential negative impacts are mitigated by the CC\,BY-NC-SA\,4.0 license,
anonymisation at capture time, and explicit guidance against surveillance
applications in the dataset card.
We encourage users to consult the dataset documentation and adhere to the
ethical guidelines outlined in Appendix~\ref{app:ethics} before deployment
in any downstream application.

\end{document}